\documentclass[10pt,twocolumn,letterpaper]{article}
\usepackage[sectionbib]{chapterbib}

\usepackage{cvpr}              
\usepackage{float}
\usepackage{array}
\usepackage{multirow}
\usepackage{diagbox}
\usepackage{makecell}
\usepackage{tabularx}
\usepackage[sectionbib]{chapterbib}
\usepackage{graphicx, balance}
\usepackage{afterpage}
\usepackage{stfloats}
\usepackage{flushend}

\definecolor{cvprblue}{rgb}{0.21,0.49,0.74}
\usepackage[pagebackref,breaklinks,colorlinks]{hyperref}

\hypersetup{
    urlcolor=pink!180,         
}

\def\confName{CVPR}
\def\confYear{2025}

\title{One is Plenty: A Polymorphic Feature Interpreter for Immutable Heterogeneous Collaborative Perception}

\author{Yuchen~Xia$^{1}$ \quad Quan~Yuan$^{1}$\thanks{Corresponding author.} \quad Guiyang~Luo$^{1}$ \quad Xiaoyuan~Fu$^{1*}$  \\ 
Yang~Li$^{1}$ \quad Xuanhan~Zhu$^{1}$ \quad Tianyou~Luo$^{1}$ \quad Siheng~Chen$^{2}$ \quad Jinglin~Li$^{1}$ \\
$^{1}${State Key Laboratory of Networking and Switching Technology, } \\
Beijing University of Posts and Telecommunications \\
$^{2}${Cooperative Medianet Innovation Center, Shanghai Jiao Tong University}\\ 
$^{1}${\tt\small \{yuchen.xia, yuanquan, luoguiyang, fuxiaoyuan, leeyang866, xuanhan.zhu, }\\ 
 {\tt\small lty349, jlli\}@bupt.edu.cn}  \quad  \quad \quad 
$^{2}${\tt\small \{sihengc\}@sjtu.edu.cn} \\
}

\begin{document}
\maketitle
\begin{abstract}

Collaborative perception in autonomous driving significantly enhances the perception capabilities of individual agents. Immutable heterogeneity, where agents have different and fixed perception networks, presents a major challenge due to the semantic gap in exchanged intermediate features without modifying the perception networks. Most existing methods bridge the semantic gap through interpreters. However, they either require training a new interpreter for each new agent type, limiting extensibility, or rely on a two-stage interpretation via an intermediate standardized semantic space, causing cumulative semantic loss. To achieve both extensibility in immutable heterogeneous scenarios and low-loss feature interpretation, we propose PolyInter, a polymorphic feature interpreter. It provides an extension point where new agents integrate by overriding only their specific prompts, which are learnable parameters that guide interpretation, while reusing PolyInter's remaining parameters. By leveraging polymorphism, our design enables a single interpreter to accommodate diverse agents and interpret their features into the ego agent's semantic space. Experiments on the OPV2V dataset demonstrate that PolyInter improves collaborative perception precision by up to 11.1\% compared to SOTA interpreters, while comparable results can be achieved by training only 1.4\% of PolyInter's parameters when adapting to new agents.
Code is available at \href{https://github.com/yuchen-xia/PolyInter}{https://github.com/yuchen-xia/PolyInter}.

\end{abstract}

\section{Introduction}
\label{sec:intro}

\begin{figure*}[t]
  \centering
   \includegraphics[width=0.9\linewidth]{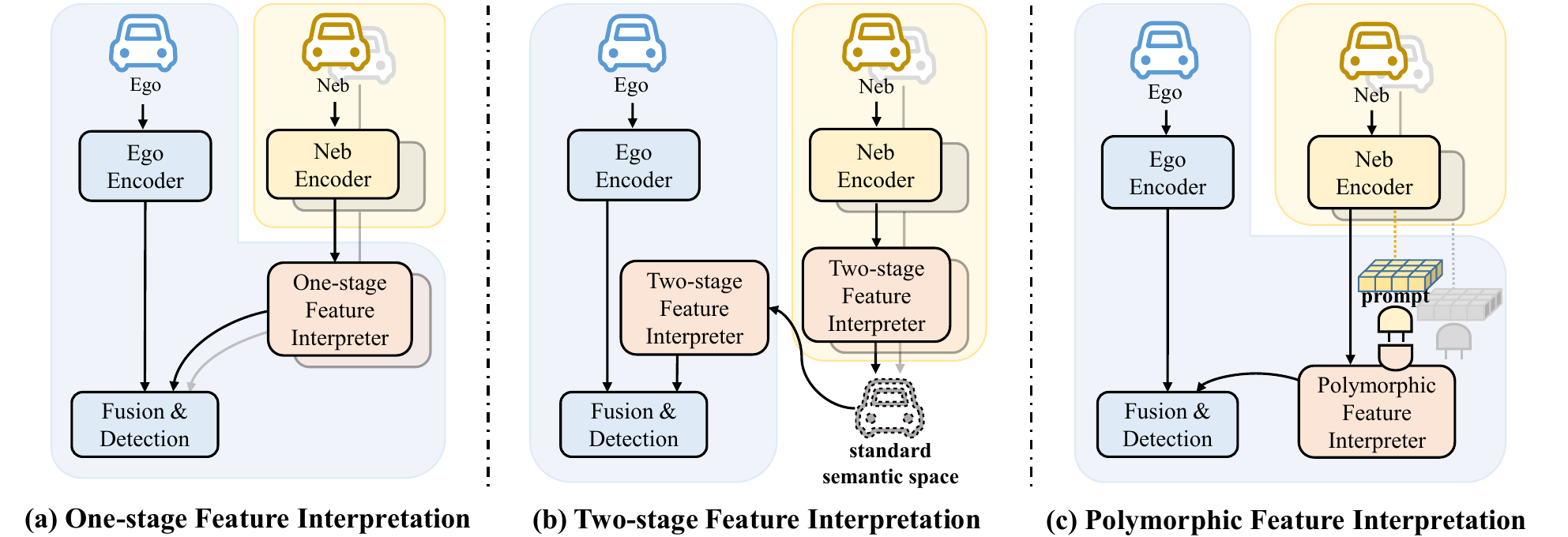}

   \caption{Comparison of  different immutable heterogeneous collaborative strategies for extending collaboration with new neighbor agents. (a) interprets neighbor features directly into the ego agent’s semantic space in a one-stage interpretation. (b) requires a two-stage feature interpretation for each collaboration, using a standard semantic space. (c) leverages a polymorphic feature interpreter, requiring only prompt tuning for each new neighbor agent. The blue areas are on the ego agent, while the yellow areas are on the neighbor agents.}
   \label{motivation}
\end{figure*}

Collaborative perception is essential for driving safety of autonomous vehicles \cite{DBLP:journals/tits/ArnoldDTF22,liuWhen2comMultiAgentPerception2020}. It aims to improve each agent’s understanding of driving environment through perceptual data exchanging, which addresses challenges like occlusion and adverse weather conditions. Collaborative perception has been widely applied in object detection, obstacle recognition and occupancy prediction \cite{DBLP:journals/tits/WangZG23}.

Homogeneous perception networks are typically exploited by collaborative perception, enabling the exchange of feature-level perceptual data that are aligned in both semantics and granularity \cite{DBLP:journals/corr/abs-2403-11371,luRobustCollaborative3D2023,huWhere2commCommunicationEfficientCollaborative2022a}. However, perception networks differ significantly across vehicles of different models or from different manufacturers,  leading to transmitted intermediate features that are too heterogeneous to be understood by other agents. Furthermore, perception networks are critical to driving safety and are tightly coupled with downstream tasks, making them difficult to replace or retrain. These limitations give rise to the challenge of \textbf{immutable heterogeneous collaborative perception}.




Some pioneering works \cite{xuBridgingDomainGap2023,luoPlugPlayRepresentation} have explored the problem of immutable heterogeneous collaborative perception by focusing on resolving differences in semantics and feature size while preserving the original perception networks. In these works, immutable heterogeneity is often addressed using interpreters, which primarily perform semantic interpretation, along with feature size transformation. However, addressing the openness in immutable heterogeneous collaboration remains a challenging issue, as it requires \emph{an interpreter to efficiently extend to emerging agents without incurring substantial retraining costs or storage overheads for neural networks.} Existing interpreter-based methods generally follow two strategies: 1) One-stage strategy \cite{xuBridgingDomainGap2023}, as shown in Figure \ref{motivation}(a), requires training and storing a separate interpreter on the ego agent to align neighbor features in semantics with ego features. However, this strategy fails to address the openness of immutable heterogeneous collaborative perception, as it requires training a separate interpreter for each new type of collaborating neighbor agent. 2) Two-stage strategy \cite{luoPlugPlayRepresentation}, as shown in Figure \ref{motivation}(b), defines a standard semantic space to enhance extensibility, where each agent only needs to train and store two interpreters to interpret its own semantics from/to the standard semantic space. However, this approach requires two stages of interpretation, leading to cumulative semantic loss at each stage.

To achieve both high extensibility and low semantic loss in immutable heterogeneous collaboration, we propose PolyInter, a Polymorphic Feature Interpreter, as illustrated in Figure \ref{motivation}(c). Polymorphism means that the interpreter can exhibit different behaviors based on the input. To implement polymorphism, the generation of PolyInter consists of two phases: base model training and generalization. In the first phase, existing agents jointly train the interpreter on the ego agent, which consists of an interpreter network, a general prompt, and a specific prompt for each neighbor agent. Here, the interpreter network interprets the semantics of the neighbor agents into the ego agent’s semantic space, while the prompts, similar to visual prompts \cite{jiaVisualPromptTuning2022}, are learnable parameters guiding the feature interpretation. In the second phase, PolyInter extends collaboration with new neighbor agents by ``inheriting" the previously trained interpreter, overriding and fine-tuning \cite{DBLP:conf/cvpr/Ma0Y023,DBLP:conf/cvpr/BaiZGLGHHL24} the specific prompts for the new neighbors, while keeping the parameters of the interpreter network and the general prompt frozen. Consequently, to collaborate with various neighbor agents, the ego agent needs only to store a single interpreter equipped with tailored prompts for each neighbor, enabling one-stage interpretation of features across different semantic spaces and alignment with the ego agent’s semantic space. This design enhances extensibility and minimizes semantic loss.

In PolyInter, to allow the shared interpreter network to handle differences in the channel and spatial distribution of heterogeneous semantics, which are received from different neighbor agents, a Channel Selection Module is introduced for semantic alignment along the channel dimension, and a Spatial Attention Module is designed for spatial semantic correlation. Additionally, to extract shared semantics from heterogeneous features while interpreting agent-specific semantic information, both general and specific prompts are designed. Average precision (AP) at Intersection-over-Union (IoU) thresholds of 0.5 and 0.7 is adopted to evaluate perception performance. Our key contributions are threefold: 



\begin{itemize}
    \item We design a highly extensible polymorphic feature interpreter for immutable heterogeneous collaborative perception, enabling the ego agent to perform one-stage semantic interpretation with new emerging neighbor agent types while storing only a single interpreter network and a set of prompts, with minimal parameter fine-tuning needed.
\end{itemize}
\begin{itemize}
    \item PolyInter achieves holistic semantic alignment by addressing differences in channel and spatial distributions through the channel selection and spatial attention modules, while using learnable prompts to handle semantic encoding variations.
\end{itemize}
\begin{itemize}
    \item Comprehensive experiments on OPV2V dataset show that PolyInter improves AP by 7.9\% and 11.1\% at IoU thresholds of 0.5 and 0.7, respectively, compared to SOTA immutable heterogeneous feature interpreters.
\end{itemize}

\section{Related Work}
\label{sec:relatedwork}

\subsection{Collaborative Perception}
Collaborative perception enhances precision by utilizing shared perceptual data from multiple agents, typically categorized into early, intermediate, and late fusion techniques. Among these, intermediate fusion \cite{liuVehicleeverythingAutonomousDriving2023, hanCollaborativePerceptionAutonomous2023,xuCoBEVTCooperativeBirds2022} is preferred for its optimal trade-off between performance and bandwidth efficiency.
To address communication overhead, Hu et al. \cite{huWhere2commCommunicationEfficientCollaborative2022a,DBLP:conf/cvpr/HuPLGLC24} proposed Where2comm and CodeFilling, which enhance communication efficiency by reducing redundancy. Liu et al. \cite{liuWhen2comMultiAgentPerception2020} proposed a framework that learns when and with whom to communicate. Luo et al. \cite{luoComplementarityEnhancedRedundancyMinimizedCollaboration2022} developed CRCNet to minimize redundancy among shared features. Some methods \cite{leiLatencyAwareCollaborativePerception2022,renInterruptionAwareCooperativePerception2024,Yang_2023_ICCV} aimed to address communication interruption or latency issues by leveraging techniques such as historical cooperation information.


\subsection{Heterogeneous Feature Fusion}
Recent research has addressed the heterogeneity between agents to improve the effectiveness of collaborative perception. Some methods \cite{zhaoBM2CPEfficientCollaborative2023,xiangHMViTHeteromodalVehicleVehicle2023,10265751} mainly handled the heterogeneity between LiDAR and camera modalities by designing specialized network architectures. HEAL \cite{luExtensibleFrameworkOpen2024}, V2X-ViT \cite{xuV2XViTVehicleEverythingCooperative2022} and Hetecooper \cite{shaoHetecooperFeatureCollaboration} designed various feature transformation methods or collaborative networks to address the issue of sensor heterogeneity. Additionally, MPDA \cite{xuBridgingDomainGap2023} and PnPDA \cite{luoPlugPlayRepresentation} bridged the domain gap in multi-agent perception by using a feature interpreter to align the heterogeneous semantics of neighbor agents with the ego agent’s semantic space, thereby tackling the challenge of immutable heterogeneous collaborative perception. Building on this foundation, PolyInter aims to generalize to new agents using only a single interpreter network.

\subsection{Visual Prompt}
Visual prompting has emerged as an innovative technique for enhancing model performance across various domains by incorporating additional guiding visual information. Jia et al. \cite{jiaVisualPromptTuning2022} introduced visual prompt tuning, an efficient way to adjust vision Transformers without full fine-tuning, by adding trainable prompts to the input space while keeping the model backbone frozen. Bar et al. \cite{barVisualPromptingImage2022} designed prompts tailored to specific tasks. In addition, methods for multi-source domain adaptation \cite{chenMultiPromptAlignmentMultiSource2023,ganDecorateNewcomersVisual2023,singhaADCLIPAdaptingDomains2023,yangExploringSparseVisual2024,caoDomainControlledPromptLearning2024} aimed to align features across domains, enhancing generalization by adapting the model to multiple sources simultaneously. Other methods \cite{liLearningDomainAwareDetection2023,chengDisentangledPromptRepresentation2024,caiViPLLaVAMakingLarge2024} focused specifically on improving prompt representations to boost generalization and adaptability across both domains and tasks. These include techniques such as domain-aware prompt tuning for object detection under domain shifts, disentangled prompt representations for domain generalization, and multi-modal prompt models for interpreting diverse visual prompts across tasks.


\section{Methodology}
\label{sec:methodology}

\begin{figure*}[t]
  \centering
   \includegraphics[width=0.95\linewidth]{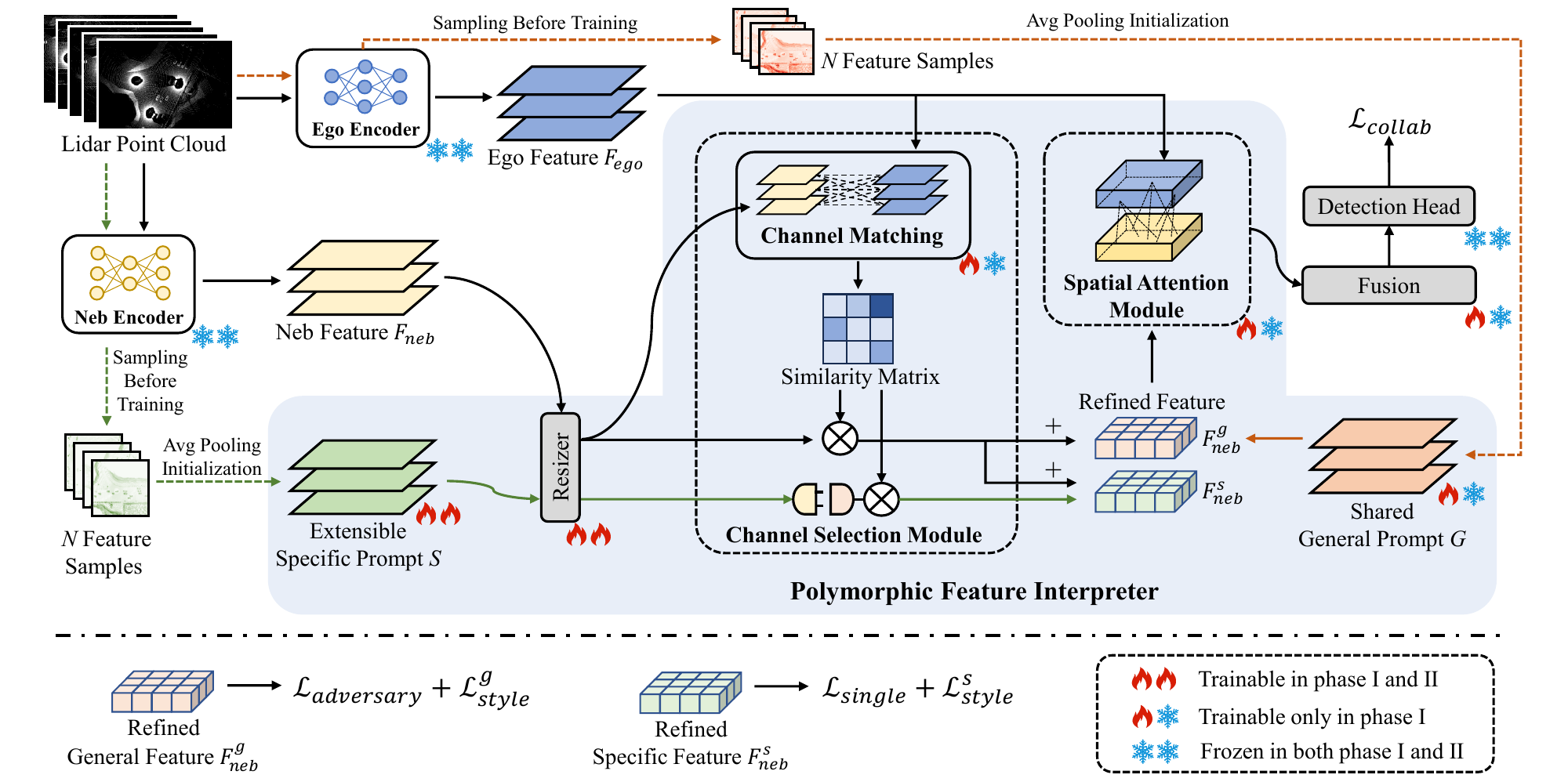}

   \caption{The overall architecture of PolyInter. PolyInter establishes a common structure that can be inherited by multiple agents, providing an extension point for customizing the specific prompts of each agent. This interpreter incorporates both a Channel Selection Module and a Spatial Attention Module to facilitate feature semantic interpretation. }
   \label{framework}
\end{figure*}

Figure \ref{framework} shows the overall architecture of PolyInter, with each module's parameters marked as either trainable or frozen during Phase \uppercase\expandafter{\romannumeral1} and Phase \uppercase\expandafter{\romannumeral2}. 

\subsection{Overall Pipeline}
This paper focuses on immutable heterogeneous scenarios where agents have different perception network structures (including fixed encoders and detection heads), causing significant variations in the semantics and granularity of the features transmitted between them. To align the semantics of neighbor features with those of the ego features, PolyInter on the ego agent includes a unified interpreter network, a shared general prompt, and multiple specific prompts corresponding to each neighbor agent. The interpreter network consists of a channel selection module for channel-level alignment and a spatial attention module for spatial-level alignment. The training process is divided into two phases: Phase \uppercase\expandafter{\romannumeral1} is for base model training, and Phase \uppercase\expandafter{\romannumeral2} is for generalization.


Before the training begins, a general prompt $G$ needs to be initialized to extract shared semantic information between the agents, and a specific prompt $S_{i}$ should be initialized for each neighbor agent $i$ to extract the agent-specific semantic information. 

In training Phase \uppercase\expandafter{\romannumeral1}, the observation data from $n$ collaborating neighbor agents $\{O_{1},O_{2},...,O_{n}\}$ is input into the encoders of the $n$ agents $\{\mathrm{ENC}_{1}(),\mathrm{ENC}_{2}(),...,\mathrm{ENC}_{n}() \}$, obtaining a set of heterogeneous intermediate features $\{ F_{neb,1},F_{neb,2},...,F_{neb,n} \}$. PolyInter on the ego agent takes these heterogeneous features as inputs, learning the interpreter network, the shared general prompt $G$ and the extensible specific prompts $S_{i}$ corresponding to each of the neighbor agents $i$.

In training Phase \uppercase\expandafter{\romannumeral2}, PolyInter on the ego agent no longer requires training and adapts to new neighbor agents through its extension point, which enables seamless integration by fine-tuning only specific prompts associated with the new neighbor agents, leaving the core interpreter unchanged. Specifically, for each new neighbor agent $i$, a specific prompt $S_{i}$ is initialized, and only this specific prompt and a necessary resizer are updated during training. This approach utilizes the shared semantic information between heterogeneous agents, while also incorporating the unique semantic information specific to the new neighbor agent.

\subsection{Polymorphic Feature Interpreter}
\begin{figure}
  \centering
   \includegraphics[width=0.80\linewidth]{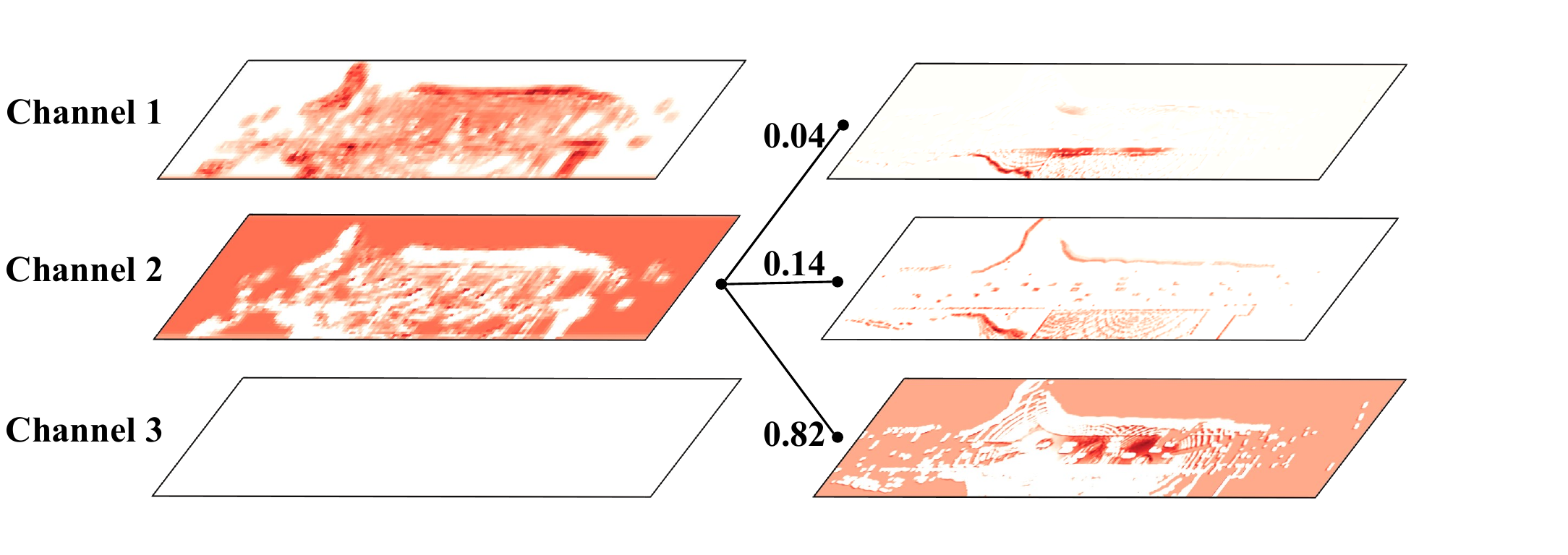}

   \caption{Visualization of BEV feature maps from two heterogeneous encoders along the corresponding channel dimension: PointPillar \cite{DBLP:conf/cvpr/LangVCZYB19} on the left and VoxelNet \cite{DBLP:conf/cvpr/ZhouT18} on the right. The numbers in the middle represent channel-wise similarity.}
   \label{channel}
\end{figure}

\textbf{Channel selection module.}
Each channel of a feature represents a dimension of semantic information. To interpret feature semantics, the first step is to align the two features along the channel dimension. As shown in Figure \ref{channel}, the Bird’s Eye View (BEV) features from two heterogeneous encoders cannot be directly aligned along the channel dimension. Consequently, it is necessary to identify the corresponding semantic components in the neighbor features for each channel of the ego feature. Existing methods \cite{CBAM} focus solely on computing attention scores within a single feature's channels, making them unsuitable for scenarios requiring cross-channel matching between distinct features. Thus, a channel selection module is designed to allow the ego feature to select and weight the neighbor feature's channels based on similarity scores. 


Specifically, the channel selection module uses cosine similarity to match the channels of the ego feature $F_{ego}$ and the currently collaborating neighbor feature $F_{neb}$ (with the index $i$ omitted in cases where no ambiguity exists), generating a similarity matrix $M\in \mathbb{R}^{C_{1}\times C_{2}}$, where $C_{1}$ and $C_{2}$ refer to the channel dimension size of $F_{ego}$ and $F_{neb}$, respectively. Each element $M[a,b]$ denotes the similarity score between the $a$-th channel of $F_{ego}$ and the $b$-th channel of $F_{neb}$, with scores ranging from 0 to 1. The following equations define this process:
\begin{equation}
\label{QK}
Q = {F}_{ego}' W_{q}, \quad K = {F}_{neb}' W_{k}, 
\end{equation}
\begin{equation}
\label{sim_mat}
M = \mathrm{SoftMax}(\frac{QK^{\top}}{\sqrt{d_{k}}}) ,
\end{equation}
where $F' \in \mathbb{R}^{C\times HW}$ is obtained by flattening the last two dimensions of $F \in \mathbb{R}^{C\times H \times W}$, and $W_{q}$ and $W_{k}$ are fully-connected layer weights. $C$, $H$, and $W$ denote the size of feature’s channels, height, and width, respectively. $d_k$ represents the dimensionality of $Q$ and $K$, used to scale the input values of $\mathrm{SoftMax}$ to an appropriate range. Both the neighbor feature $F_{neb}$ and the specific prompt $S_{i}$, corresponding to the neighbor agent are then reorganized as:
\begin{equation}
\label{F_rearrange}
F_{neb}^{r} = \mathrm{LN}(M F_{neb}), \quad S_{i}^{r} = \mathrm{LN}(M S_{i}),
\end{equation}
where $F_{neb}^{r}$ and $S_{i}^{r}$ represent the reorganized neighbor feature and specific prompt along the channel dimension, respectively, and $\mathrm{LN}$ denotes layer normalization. The general prompt $G$ and the reorganized specific prompt $S_{i}^{r}$ are then added to $F_{neb}^{r}$, resulting in the refined general feature $F_{neb}^{g}$ and refined specific feature $F_{neb}^{s}$:
\begin{equation}
\label{GS_ref}
F_{neb}^{g} = F_{neb}^{r} + G, \quad F_{neb}^{s} = F_{neb}^{r} + S_{i}^{r}, 
\end{equation}
which are used in the subsequent steps to learn shared semantic information and agent-specific semantic information through the loss functions, respectively.

\noindent\textbf{Spatial attention module.} In addition to aligning the channels and reorganizing the semantic components of the features, the process of semantics interpretation also requires learning the semantic relationships between different positions within the BEV features. To align the ego feature and neighbor feature in the spatial dimensions, a spatial attention module is employed. This module leverages a 3D attention mechanism, known as fused axial attention \cite{xuCoBEVTCooperativeBirds2022}. It computes both local and global spatial similarities between  the ego feature $F_{ego}$ and the refined neighbor feature, which is composed of $F_{neb}^{g}$ and $F_{neb}^{s}$. Through this process, the module maps the spatial semantic information of the refined neighbor feature into the semantic space of the ego feature, ensuring precise spatial-level feature interpretation. 

\noindent\textbf{Prompts initialization.} As learnable parameters, prompts are designed with two initialization methods. Sampling-based initialization incorporates statistical characteristics of heterogeneous feature distributions to improve perception, while random initialization enables further reduction in trainable parameters via low-rank decomposition.

For sampling-based initialization, both the general prompt $G$ and the specific prompt $S_{i}$ are initialized by sampling from the training dataset. Random sampling is performed on the training set to obtain $N$ point cloud samples. For the general prompt $G$, these $N$ samples are fed into the ego agent's encoder, and the resulting features are aggregated via average pooling. For the specific prompt $S_{i}$, $N$ samples are processed through the encoder of the collaborating neighbor agent and pooled to generate the corresponding specific prompt. To ensure compatibility during fusion, the sizes of both $G$ and $S_{i}$ are aligned with the ego feature, except for the channel dimension of $S_{i}$, which initially matches that of the neighbor feature and is later resized to match the ego feature's channel dimension through a 1D convolution in the resizer.

For random initialization, to reduce the number of trainable parameters, matrix low-rank decomposition \cite{DBLP:conf/iclr/HuSWALWWC22} is applied to the specific prompt $S_{i}$, as it constitutes the primary trainable parameters in the generalization phase. The original size of $S_{i}$ is $C\times H \times W$. On one hand, we apply low-rank decomposition along the last two dimensions, factorizing the $H\times W$ matrix into the product of $H\times R$ and $R\times W$ matrices ($R \ll H \text{ and } W$), reducing the parameter count from $C\times H \times W$ to $C\times R \times(H+W)$. On the other hand, we reduce the depth of the specific prompt by scaling down $C$ by a factor of $T$ when $R$ is sufficiently small, further minimizing the trainable parameters.



\subsection{Phase \uppercase\expandafter{\romannumeral1}: Base Model Training}
In the training Phase \uppercase\expandafter{\romannumeral1}, our goal is to obtain a well-generalized polymorphic feature interpreter on the ego agent that can provide an extension point for newly added agents in the next training phase. To achieve this goal, the information in the features is decoupled into shared-semantic information and agent-specific semantic information. The general prompt and specific prompt are tasked with learning these two aspects, respectively. However, solely relying on end-to-end training with the collaborative perception loss function is insufficient and may result in overfitting to the encoders of the neighbor agents used during Phase \uppercase\expandafter{\romannumeral1} training. Therefore, the style loss and the adversarial loss are employed to capture shared and agent-specific semantic information, which are used to generate the refined specific feature $F_{neb}^{s}$ and refined general feature $F_{neb}^{g}$, as output by the channel selection module in (\ref{GS_ref}). During the training process, all encoders and the ego agent's detection head are frozen, with only the parameters of the feature interpreter, PolyInter, being trained.



\noindent\textbf{Learning agent-specific semantics.} 
To ensure $F_{neb}^{s}$ captures as much agent-specific semantic information as possible, it is designed to excel in single-agent object detection. The single-agent loss $\mathcal{L}_{single}$ encourages PolyInter to learn and retain the specific semantic meanings encoded by the unique encoder of each neighbor agent. This allows the specific prompt to effectively utilize these features for independent object detection tasks, even in the absence of collaborative input:
\begin{equation}
\label{single}
\mathcal{L}_{single} = \mathcal{L}_{s\_box} + \mathcal{L}_{s\_reg},
\end{equation}
where $\mathcal{L}_{s\_box}$ and $\mathcal{L}_{s\_reg}$ represent the bounding box classification and regression losses of a single agent, respectively. 

Additionally, a style loss for $F_{neb}^{s}$ is introduced to balance feature distributions among neighbor agents. This loss aligns the statistical distributions of features between neighbor agents and the ego agent, preventing excessive dependence on any single agent:
\begin{equation}
\begin{aligned}
\label{style}
\mathcal{L}_{style}^{s} = ||\mu(F_{neb}^{s}) - \mu(F_{ego})||_{2}     + ||\sigma(F_{neb}^{s}) - \sigma(F_{ego})||_{2}
\end{aligned},
\end{equation}
where $\mu(\cdot)$ and $\sigma(\cdot)$ denote the mean and standard deviation of features, respectively.

Thus, the constraint function for the refined prompt $F_{neb}^{s}$ can be expressed as:
\begin{equation}
\label{single}
\mathcal{L}_{s} = \mathcal{L}_{single} + \omega\mathcal{L}_{style}^{s},
\end{equation}
where $\omega$ is the balancing hyper-parameter.

\noindent\textbf{Learning shared semantics.}
The refined general feature $F_{neb}^{g}$ is responsible for extracting the shared semantic parts from the features of different agents.
To achieve this, the general prompt is trained with adversary loss. This ensures the extracted features originate exclusively from the shared semantics, making them indistinguishable to the discriminator $\mathrm{D}$ attached to $F_{neb}^{g}$. Consequently, the general prompt filters out the specific semantic parts of heterogeneous features and preserves their shared semantic parts.
The adversarial optimization objective is formulated as:
\begin{equation}
\label{adversarial}
\mathcal{L}_{adversary}=\mathop\mathrm{max}\limits_{\Phi} \mathop\mathrm{min}\limits_{D} (\mathrm{E}_{ego}(\mathrm{D}(F_{neb}^{g})) + \mathrm{E}_{neb}(\mathrm{D}(F_{neb}^{g}))),
\end{equation}
where $\mathrm{E}_{ego}(\mathrm{D}(F_{neb}^{g}))$ and $\mathrm{E}_{neb}(\mathrm{D}(F_{neb}^{g}))$ represent the classification errors when the discriminator predicts whether the feature belongs to the ego agent or the neighbor agent, respectively. $F_{neb}^{g}$ is obtained by (\ref{GS_ref}), and $\Phi$ consists of the parameters of the channel selection module and the general prompt $G$.

The style loss is also applied to the refined general feature $F_{neb}^{g}$. Thus, the constraint function for $F_{neb}^{g}$ can be formulated as:
\begin{equation}
\label{style}
\mathcal{L}_{g} = \mathcal{L}_{adversary} + \omega\mathcal{L}_{style}^{g},
\end{equation}
where $\omega$ is the balancing hyper-parameter. 

\noindent\textbf{Total loss.} In the end-to-end training of phase \uppercase\expandafter{\romannumeral1}, PolyInter's interpretation capability is enhanced through a collaborative object detection task while imposing distinct constraints on $F_{neb}^{s}$ and $F_{neb}^{g}$. The total loss can be formulated as:
\begin{equation}
\label{phase1}
\mathcal{L}_{phase \uppercase\expandafter{\romannumeral1}} =  \mathcal{L}_{collab} + \lambda_s\mathcal{L}_{s} + \lambda_g\mathcal{L}_{g},
\end{equation}
where $\mathcal{L}_{collab}$ donates the collaborative loss during the object detection performed jointly by the ego agent and the neighbor agent. Its calculation is the same as that of $\mathcal{L}_{single}$. $\lambda_s$ and $\lambda_g$ are hyper-parameters. 

\subsection{Phase \uppercase\expandafter{\romannumeral2}: Generalization}
PolyInter trained in Phase \uppercase\expandafter{\romannumeral1} is endowed with generalization capabilities, providing an extension point for newly introduced agents to integrate by leveraging their specific prompts. Consequently, in Phase \uppercase\expandafter{\romannumeral2}, instead of retraining the entire interpreter, only the extensible specific prompt $S_{i}$ associated with the new agent is optimized.

To enable the refined specific feature $F_{neb}^{s}$ to capture the agent-specific semantic information of new agent types, the same loss constraints from Phase \uppercase\expandafter{\romannumeral1} is applied to regulate $F_{neb}^{s}$. Additionally, to effectively interpret these information into the ego agent's semantic space, the learning of $F_{neb}^{s}$ is constrained using the collaborative object detection loss. Thus, the loss function for $F_{neb}^{s}$ is formulated as:
\begin{equation}
\label{phase1}
\mathcal{L}_{phase \uppercase\expandafter{\romannumeral2}} =  \mathcal{L}_{collab} + \lambda_s\mathcal{L}_{s}.
\end{equation}


\subsection{Application Paradigm}
During the inference process, the ego agent is equipped with a single PolyInter to manage all collaborative tasks. PolyInter maintains a shared general prompt, along with $N$ extensible specific prompts for $N$ collaborating agent types.

In non-collaborative scenarios, the ego agent processes its own features independently, bypassing PolyInter. When collaboration is required, neighbor agents broadcast intermediate features in their respective semantic spaces, prompting the ego agent to activate the corresponding specific prompts and resizers. PolyInter on the ego agent then performs real-time feature interpretation, aligning the received semantics with the ego feature.
\section{Experiments}
\label{sec:experiments}

\subsection{Settings}

\textbf{Dataset.}
OPV2V dataset \cite{DBLP:conf/icra/XuXXHLM22}, a large-scale public benchmark for collaborative perception in autonomous driving, is utilized in our experiments. Additionally, comparative experiments on the simulated dataset V2XSet \cite{xuV2XViTVehicleEverythingCooperative2022} and the real-world dataset DAIR-V2X \cite{DBLP:conf/cvpr/YuLSHYSGLHYN22} are included in the supplementary material due to space constraints.

\noindent\textbf{Experiment design.}
PointPillar \cite{DBLP:conf/cvpr/LangVCZYB19}, VoxelNet \cite{DBLP:conf/cvpr/ZhouT18} and SECOND \cite{DBLP:journals/sensors/YanML18} are commonly used LiDAR feature encoders, each available in multiple configurations. Table \ref{parameters} details heterogeneous encoder parameters and collaborative performance in homogeneous settings. In Phase \uppercase\expandafter{\romannumeral1}, the base model is trained with two distinct encoder combinations, represented in the format of ``ego-neb1-neb2" as: {pp8-vn4-sd2} and {pp8-pp4-vn4}. The ego agent randomly selects a neighbor encoder per batch for training. During Phase \uppercase\expandafter{\romannumeral2}, the ego agent collaborates with new neighbor agents pp4, sd1 and vn6 using the pre-trained PolyInter, training only the specific prompt and resizer for each neighbor. Collaborative perception precision is then evaluated for each two-agent scenario, in the format of ``ego-neb" as: pp8-pp4, pp8-sd1 and pp8-vn6, while three-agent collaborative experiments are included in the supplementary material. In all experiments, the encoders of all collaborating agents and the detection head of the ego agent are kept frozen and inaccessible, ensuring the evaluation of extensibility in 
open immutable heterogeneous collaboration scenarios.

In 3D collaborative object detection, the evaluation range is set to $x\in\left[-140, 140\right]$ meters and $y\in\left[-40,40\right]$ meters to ensure consistency with previous methods \cite{xuBridgingDomainGap2023,luoPlugPlayRepresentation}.

\begin{table}[t]
  \centering
  \resizebox{0.48\textwidth}{!}{
  \begin{tabular}{@{}>{\raggedright\arraybackslash}p{2.2cm}|>{\centering\arraybackslash}p{1.2cm}|>{\centering\arraybackslash}p{1.7cm}|>{\centering\arraybackslash}p{1.8cm}|>{\centering\arraybackslash}p{1.7cm}|>{\centering\arraybackslash}p{1.6cm}@{}}
    \toprule[\heavyrulewidth]
     \multirow{2}{*}{Encoder} & \multirow{2}{*}{Variation} & Voxel Resolution & 2D / 3D CNN Layers & Half Lidar Range (x,y) & AP@0.5 / AP@0.7\\
    \midrule
     \multirow{3}{*}{\parbox[c]{2.2cm}{PointPillar \cite{DBLP:conf/cvpr/LangVCZYB19}}} & pp8 & 0.8, 0.8, 4 & 10 / 0 & 140.8, 40 & 83.9 / 69.2 \\
     & pp6 & 0.6, 0.6, 4 & 19 / 0 & 153.6, 38.4 & 86.5 / 78.7 \\
     & pp4 & 0.4, 0.4, 4 & 19 / 0 & 140.8, 40 & 87.2 / 77.7 \\
     \midrule
     \multirow{2}{*}{\parbox[c]{2.2cm}{VoxelNet \cite{DBLP:conf/cvpr/ZhouT18}}} & vn6 & 0.6, 0.6, 0.4 & 0 / 3 & 153.6, 38.4 & 57.9 / 49.2 \\
     & vn4 & 0.4, 0.4, 0.4 & 0 / 3 & 140.8, 40 & 85.5 / 78.7 \\
     \midrule
     \multirow{2}{*}{\parbox[c]{2.2cm}{SECOND \cite{DBLP:journals/sensors/YanML18}}} & sd2 & 0.2, 0.2, 0.2 & 12 / 12 & 140.8, 40 & 64.4 / 53.0 \\
     & sd1 & 0.1, 0.1, 0.1 & 13 / 13 & 140.8, 40 & 65.1 / 52.9\\
    \bottomrule[\heavyrulewidth]
  \end{tabular}
  }
  \caption{Detailed parameters of heterogeneous encoders.}
  \label{parameters}
\end{table}

\begin{table*}
  \centering
  \resizebox{0.85\textwidth}{!}{
  \begin{tabular}{@{}>{\raggedright\arraybackslash}p{1.1cm}lcccccccc@{}}
    \toprule
    \multicolumn{2}{c}{\multirow{2}{*}{\diagbox[width=2.5cm,height=1cm]{Scenarios }{Interpreter}}} & \multicolumn{3}{c}{{w/ F-cooper \cite{DBLP:conf/edge/ChenMTGYF19} Fusion}}& & \multicolumn{3}{c}{{w/ CoBEVT \cite{xuCoBEVTCooperativeBirds2022} Fusion}}  \\
    \cmidrule[\heavyrulewidth]{3-5} \cmidrule[\heavyrulewidth]{7-9}
     & & \textbf{PolyInter (Ours)} & \textbf{PnPDA} \cite{luoPlugPlayRepresentation} & \textbf{MPDA} \cite{xuBridgingDomainGap2023} &   & \textbf{PolyInter (Ours)} & \textbf{PnPDA} \cite{luoPlugPlayRepresentation} & \textbf{MPDA} \cite{xuBridgingDomainGap2023} \\
    \midrule
    \makebox[2cm][l]{pp8-pp4*}&\multirow{2}{*}{\cite{DBLP:conf/cvpr/LangVCZYB19}} & \textbf{77.6 / 60.9} & \multirow{2}{*}{77.1 / 51.1} & \multirow{2}{*}{76.5 / 53.1} &    & \textbf{80.2 / 65.4} & \multirow{2}{*}{79.2 / 62.3} & \multirow{2}{*}{76.7 / 54.9} \\
    \makebox[2cm][l]{pp8-pp4+} &  & \textbf{77.2 / 65.8} & &    & & \textbf{80.3 / 66.0} & & \\
    \midrule
    \makebox[2cm][l]{pp8-sd1*}&\multirow{2}{*}{\cite{DBLP:conf/cvpr/LangVCZYB19,DBLP:journals/sensors/YanML18}} & \textbf{81.5 / 67.9} & \multirow{2}{*}{68.4 / 40.3} & \multirow{2}{*}{59.4 / 43.4} &    & \textbf{83.4 / 73.0} & \multirow{2}{*}{78.7 / 64.6} & \multirow{2}{*}{80.9 / \textbf{71.6}} \\
    \makebox[2cm][l]{pp8-sd1+} &  & \textbf{79.4 / 66.3} & &    & & \textbf{81.7} / 69.7 & & \\
    \midrule
    \makebox[2cm][l]{pp8-vn6*}&\multirow{2}{*}{\cite{DBLP:conf/cvpr/LangVCZYB19,DBLP:conf/cvpr/ZhouT18}} & \textbf{71.9 / 51.0} & \multirow{2}{*}{61.9 / 44.2} & \multirow{2}{*}{58.9 / 43.7} &    & \textbf{73.2 / 56.7} & \multirow{2}{*}{63.1 / 49.9} & \multirow{2}{*}{57.4 / 42.3} \\
    \makebox[2cm][l]{pp8-vn6+} &  & \textbf{72.8 / 55.1} & &    & & \textbf{73.6 / 56.3} & & \\
    
    \bottomrule
  \end{tabular}
  }
  \caption{Comparison with PnPDA and MPDA in three heterogeneous scenarios for collaborative perception, using AP@0.5/AP@0.7. PolyInter was pre-trained with two different combinations (in the format of ``ego-neb1-neb2"): Symbol ``*" indicates the pp8-vn4-sd2 combination, while symbol ``+" indicates the pp8-pp4-vn4 combination. ``Scenarios" refers to different combinations of the ego agent and the new neighbor agent in Phase \uppercase\expandafter{\romannumeral2}. Our experiments include three scenarios (in the format of ``ego-neb"): pp8-pp4, pp8-sd1, and pp8-vn6.}
  \label{main-results}
\end{table*}

\noindent\textbf{Implementation details.}
When pp8 serves as the ego agent, the general prompt is configured to match pp8's feature dimensions, with a channel count of 256, a height of 50, and a width of 176. For the neighbor agent's specific prompt, the channel dimension aligns with its feature, while height and width match the ego agent. The hyper-parameters $\mathcal{\omega}$, $\lambda_s$, $\lambda_g$ are set to 0.5, 1 and 1, respectively.

\subsection{Performance Comparison}
\label{subsec:Performance}

We compared PolyInter with PnPDA \cite{luoPlugPlayRepresentation} and MPDA \cite{xuBridgingDomainGap2023}, the only two interpreters designed for immutable heterogeneous collaboration, as shown in Table \ref{main-results}. The results for PnPDA presented here are from one-stage interpretation, which outperforms the two-stage interpretation as demonstrated in \cite{luoPlugPlayRepresentation}. Two commonly applied fusion methods F-cooper \cite{DBLP:conf/edge/ChenMTGYF19} and CoBEVT \cite{xuCoBEVTCooperativeBirds2022} are used for the fusion module. The experimental results for PolyInter represent the performance in the generalization phase, where the specific prompts of new neighbors are initialized through sampling, and only the specific prompts are fine-tuned based on the interpreter and fusion module trained in the first phase. In contrast, the results for PnPDA and MPDA reflect training all interpreter and fusion module's parameters specifically for each ego-neighbor pair.

\begin{figure}[!t]
  \centering
   \includegraphics[width=1.0\linewidth]{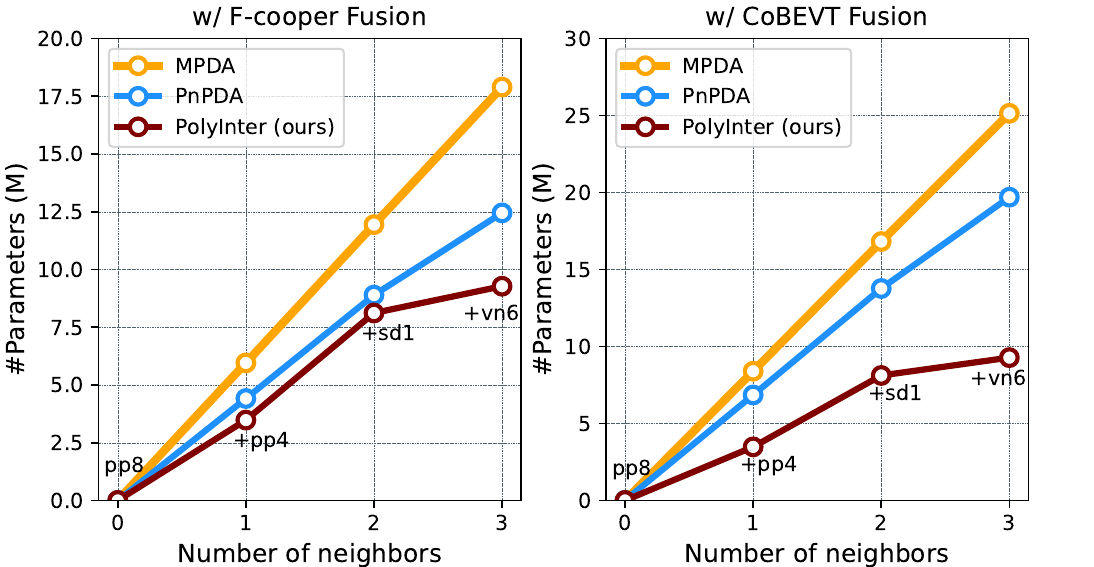}

   \caption{Comparison of the number of trainable parameters with PnPDA and MPDA when incrementally adding new heterogeneous neighbor agents.}
   \label{param_cars}
\end{figure}

In three heterogeneous collaborative perception scenarios, PolyInter achieved superior or comparable performance to other feature interpreters. In the first scenario, the heterogeneity between pp8 and pp4 lies in differences in voxel size and network parameter scale, with similar but not identical semantics. PolyInter demonstrates a 9.2\% improvement in AP at an IoU threshold of 0.7 compared to the other two interpreters, while achieving comparable performance at an IoU threshold of 0.5. In the second scenario, pp8-sd1, the encoder structures of the ego agent and the neighbor agent differ, resulting in greater semantic heterogeneity. In this case, PolyInter outperforms the other two by an average of 9.7\% and 14.3\% in terms of AP at IoU thresholds of 0.5 and 0.7, respectively. In the last scenario, pp8-vn6, the differences in encoder structure are further amplified, and PolyInter shows significant improvement, achieving an average increase of 12.6\% and 9.8\% in terms of AP at IoU thresholds of 0.5 and 0.7, respectively. This demonstrates that PolyInter can maintain strong performance even with substantial semantic differences.

To validate the extensibility of PolyInter, we compared the number of parameters required for training with those of PnPDA and MPDA when adapting to new neighbor agents, as shown in Figure \ref{param_cars}. With pp8 as the ego agent, neighbor agent types pp4, sd1, and vn6 are incrementally added for collaborative perception. Notably, PolyInter requires fewer trainable parameters than PnPDA and MPDA, as it only trains a specific prompt and resizer for each new neighbor agent type, making it highly extensible in immutable heterogeneous collaboration scenarios. PolyInter’s advantage is even more pronounced with the CoBEVT fusion method, where PnPDA and MPDA require a new fusion network for each additional neighbor agent type, causing a rapid increase in parameters and hindering extensibility.

\begin{table}[t]
  \centering
  \renewcommand\arraystretch{1.1}
  \resizebox{0.48\textwidth}{!}{
  \begin{tabular}{@{}lccc@{}}
    \toprule
     & \textbf{pp8-pp4} & \textbf{pp8-sd1} & \textbf{pp8-vn6} \\
    \midrule[\heavyrulewidth]
    PolyInter & \underline{77.2} / \textbf{65.8} & \textbf{79.4 / 66.3} & \textbf{72.8 / 55.1} \\
    \quad -w/o channel selection & 75.6 / 59.9 & 78.1 / 65.8 & 64.1 / 43.4 \\
    \quad -w/o spatial attention & 64.4 / 56.1 & 65.9 / 59.0 & 62.0 / 50.0 \\
    \quad -w/o general prompt & \textbf{78.3} / 64.0 & 78.0 / 66.1 & 70.9 / 52.9 \\
    \quad -w/o specific prompt & 23.3 / 7.8 & 38.7 / 16.8 & 14.3 / 2.3 \\
    \quad -w/o prompt init & 76.8 / 64.2 & 78.2 / 63.3 & 68.7 / 53.6 \\
    \bottomrule[\heavyrulewidth]
  \end{tabular}
  }
  \caption{Evaluation of the effectiveness of interpreter components and prompt initialization method. The fusion method used in this experiment is F-cooper. ``w/o prompt init" indicates that both the specific prompt and the general prompt are initialized randomly instead of using sampling-based initialization.}
  \label{tab:ablation}
\end{table}

\begin{figure}[!t]
  \centering
   \includegraphics[width=1.0\linewidth]{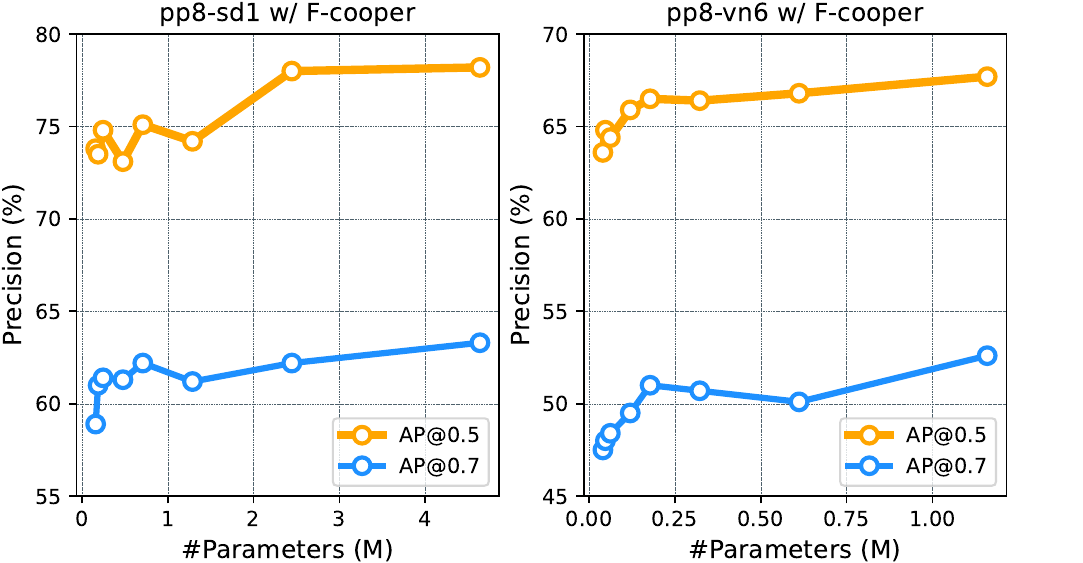}

   \caption{Experiment on the number of trainable parameters. The points on the plot from left to right correspond to $R = 1$ with $T$ values of 4, 2, and 1, followed by $T = 1$ with $R$ values of 3, 5, 10, and 20. The rightmost point represents the case where the specific prompt parameters are not decomposed.}
   \label{param_fig}
\end{figure}

\subsection{Ablation Study}

\textbf{Evaluation on main components.} Table \ref{tab:ablation} demonstrates the effectiveness of channel selection module, spatial attention module, the general prompt, and sampling-based initialization method for both prompts. PolyInter is trained in Phase \uppercase\expandafter{\romannumeral1} with pp8-pp4-vn4. In the generalization phase, the ego agent pp8 is equipped with PolyInter, while pp4, sd1, and vn6 are introduced as neighbor agents, respectively. In both phases, we individually remove key components from the interpreter and replace the prompt initialization method with random initialization to observe how these changes affect perception precision in Phase \uppercase\expandafter{\romannumeral2}.

The absence of any module or prompt decreases perception performance. The channel selection module has the most impact with vn6 as the neighbor, reducing AP by 8.1\% and 11.7\% at IoU 0.5 and 0.7, respectively, due to significant semantic differences along the channel dimension between pp8 and vn6 features. Removing the spatial attention module decreases AP by 12.8\%, 13.5\%, and 10.8\% at IoU 0.5, and 9.7\%, 7.3\%, and 5.1\% at IoU 0.7 across all heterogeneous scenarios, highlighting the need for spatial alignment after channel alignment. Without the general prompt, performance slightly declines in most scenarios, showing its role in providing shared semantics. Removing the specific prompt results in a substantial performance drop, emphasizing its importance for generalization. Finally, replacing the initialization method for both prompts slightly decreases performance, especially with vn6 as the neighbor, indicating that sampling-based prompt initialization provides valuable, customized knowledge for subsequent training.

\noindent\textbf{Decomposition of the specific prompt.}
To reduce the number of trainable parameters and lower the overhead for PolyInter during practical deployment, we apply low-rank decomposition to the specific prompt, decreasing its parameters from $C\times H \times W$ to $C\times R \times(H+W)$. When $R$ is reduced to 1, we further scale down $C$ by a factor of $T$, adjusting $T$ and $R$ to control the parameter count.

As shown in Figure \ref{param_fig}, with pp8 as the ego agent, collaborative perception performance improves as trainable parameters increase; however, the rate of improvement levels off beyond a certain threshold. Even with low trainable parameter numbers (0.16M for sd1 as the neighbor and 0.04M for vn6, accounting for only 1.4\% and 0.4\% of the respective interpreters), collaborative perception precision surpasses that of MPDA and PnPDA, demonstrating the adaptability benefits provided by the prompting mechanism.

\begin{figure}[!t]
  \centering
   \includegraphics[width=1\linewidth]{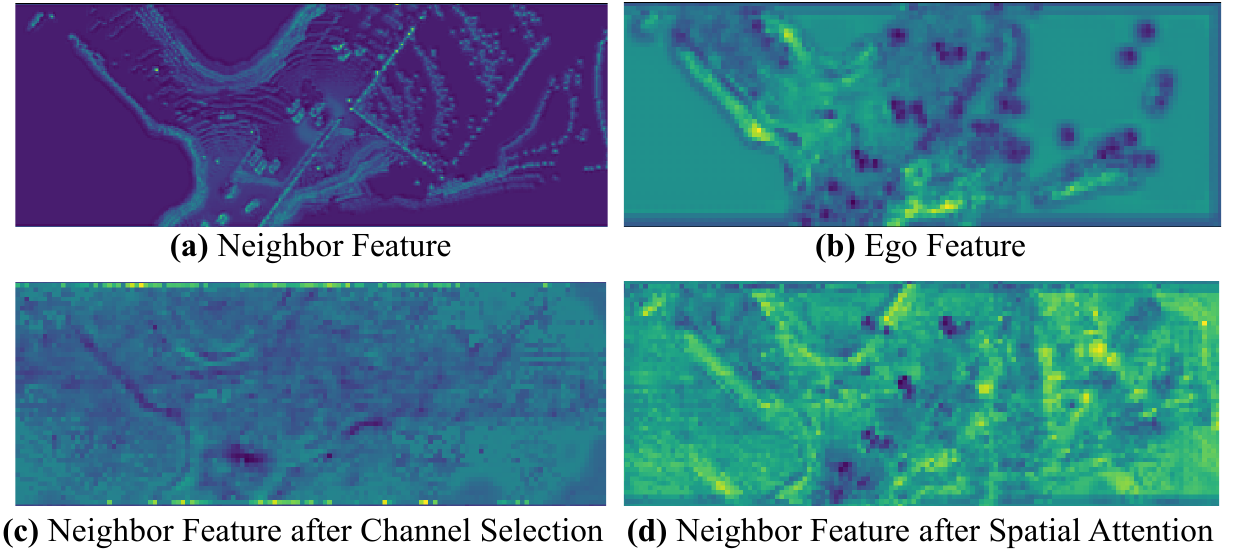}

   \caption{Visualization of intermediate features before and after interpretation. The ego agent uses pp8 encoder, and the neighbor agent uses vn4 encoder.}
   \label{visualization}
\end{figure}

\subsection{Qualitative Evaluation}
To provide a more intuitive demonstration of the effectiveness of PolyInter, we performed average pooling along the channel dimension for both features and prompts, with visualizations shown in Figure \ref{visualization}. Comparing Figure \ref{visualization}(a) and (b) shows significant semantic differences between the ego agent (pp8) and the neighbor agent (vn6), such as feature resolution and the magnitude of feature values. In Figure \ref{visualization}(c), when the neighbor feature is reorganized through the channel dimension and combined with the specific prompt, significant style transformation occurs, aligning it more closely with the ego feature, though spatial details remain partially blurred. After processing through the spatial attention module, the neighbor feature’s style further aligns with the ego feature, with spatial details like agent position more distinctly highlighted, as shown in Figure \ref{visualization}(d).


\section{Conclusion}

In this paper, we propose a polymorphic feature interpreter, PolyInter, to address the challenge of openness and improve perception precision in immutable heterogeneous collaborative perception, where perception networks vary among collaborating agents and remain fixed. PolyInter employs a single interpreter network and provides an extension point for new neighbor agents to inherit, while fine-tuning their prompts for collaboration, thereby significantly enhancing its extensibility.

\section{Acknowledgement}

This work was supported in part by the National Key Research and Development Program of China under Grant 2022YFB4300400, in part by the Natural Science Foundation of China under Grant 62272053 and Grant 62472048, in part by the Beijing Nova Program under Grant 20230484364.
{
    \small
    \bibliographystyle{ieeenat_fullname}
    \bibliography{main}

\begin{thebibliography}{44}
\providecommand{\natexlab}[1]{#1}
\providecommand{\url}[1]{\texttt{#1}}
\expandafter\ifx\csname urlstyle\endcsname\relax
  \providecommand{\doi}[1]{doi: #1}\else
  \providecommand{\doi}{doi: \begingroup \urlstyle{rm}\Url}\fi

\bibitem[Arnold et~al.(2022)Arnold, Dianati, de~Temple, and Fallah]{DBLP:journals/tits/ArnoldDTF22}
Eduardo Arnold, Mehrdad Dianati, Robert de Temple, and Saber Fallah.
\newblock Cooperative perception for 3d object detection in driving scenarios using infrastructure sensors.
\newblock \emph{{IEEE} Trans. Intell. Transp. Syst.}, 23\penalty0 (3):\penalty0 1852--1864, 2022.

\bibitem[Bai et~al.(2024)Bai, Zhang, Guo, Li, Guo, Hou, Han, and Lu]{DBLP:conf/cvpr/BaiZGLGHHL24}
Sikai Bai, Jie Zhang, Song Guo, Shuaicheng Li, Jingcai Guo, Jun Hou, Tao Han, and Xiaocheng Lu.
\newblock Diprompt: Disentangled prompt tuning for multiple latent domain generalization in federated learning.
\newblock In \emph{Proceedings of the IEEE/CVF Conference on Computer Vision and Pattern Recognition (CVPR)}, pages 27284--27293, 2024.

\bibitem[Bar et~al.(2022)Bar, Gandelsman, Darrell, Globerson, and Efros]{barVisualPromptingImage2022}
Amir Bar, Yossi Gandelsman, Trevor Darrell, Amir Globerson, and Alexei Efros.
\newblock Visual prompting via image inpainting.
\newblock In \emph{Advances in Neural Information Processing Systems (NeurIPS)}, pages 25005--25017, 2022.

\bibitem[Cai et~al.(2024)Cai, Liu, Mustikovela, Meyer, Chai, Park, and Lee]{caiViPLLaVAMakingLarge2024}
Mu Cai, Haotian Liu, Siva~Karthik Mustikovela, Gregory~P. Meyer, Yuning Chai, Dennis Park, and Yong~Jae Lee.
\newblock Vip-llava: Making large multimodal models understand arbitrary visual prompts.
\newblock In \emph{Proceedings of the IEEE/CVF Conference on Computer Vision and Pattern Recognition (CVPR)}, pages 12914--12923, 2024.

\bibitem[Cao et~al.(2024)Cao, Xu, Chen, Ma, and Yang]{caoDomainControlledPromptLearning2024}
Qinglong Cao, Zhengqin Xu, Yuntian Chen, Chao Ma, and Xiaokang Yang.
\newblock Domain-controlled prompt learning.
\newblock \emph{In AAAI Conference on Artificial Intelligence (AAAI)}, pages 936--944, 2024.

\bibitem[Chen et~al.(2023)Chen, Han, Wu, and Jiang]{chenMultiPromptAlignmentMultiSource2023}
Haoran Chen, Xintong Han, Zuxuan Wu, and Yu-Gang Jiang.
\newblock Multi-{{prompt alignment}} for {{multi-source unsupervised domain adaptation}}.
\newblock \emph{In Advances in Neural Information Processing Systems (NeurIPS)}, pages 74127--74139, 2023.

\bibitem[Chen et~al.(2019)Chen, Ma, Tang, Guo, Yang, and Fu]{DBLP:conf/edge/ChenMTGYF19}
Qi Chen, Xu Ma, Sihai Tang, Jingda Guo, Qing Yang, and Song Fu.
\newblock F-cooper: feature based cooperative perception for autonomous vehicle edge computing system using 3d point clouds.
\newblock In \emph{Proceedings of the 4th ACM/IEEE Symposium on Edge Computing (SEC)}, page 88–100, 2019.

\bibitem[Cheng et~al.(2024)Cheng, Xu, Jiang, Wang, Li, and Gao]{chengDisentangledPromptRepresentation2024}
De Cheng, Zhipeng Xu, Xinyang Jiang, Nannan Wang, Dongsheng Li, and Xinbo Gao.
\newblock Disentangled prompt representation for domain generalization.
\newblock In \emph{Proceedings of the IEEE/CVF Conference on Computer Vision and Pattern Recognition (CVPR)}, pages 23595--23604, 2024.

\bibitem[Gan et~al.(2023)Gan, Bai, Lou, Ma, Zhang, Shi, and Luo]{ganDecorateNewcomersVisual2023}
Yulu Gan, Yan Bai, Yihang Lou, Xianzheng Ma, Renrui Zhang, Nian Shi, and Lin Luo.
\newblock Decorate the {{newcomers}}: {{visual domain prompt}} for {{continual test time adaptation}}.
\newblock \emph{In AAAI Conference on Artificial Intelligence (AAAI)}, pages 7595--7603, 2023.

\bibitem[Han et~al.(2023)Han, Zhang, Li, Jin, Lang, and Li]{hanCollaborativePerceptionAutonomous2023}
Yushan Han, Hui Zhang, Huifang Li, Yi Jin, Congyan Lang, and Yidong Li.
\newblock Collaborative perception in autonomous driving: Methods, datasets, and challenges.
\newblock \emph{{IEEE} Intell. Transp. Syst. Mag.}, 15\penalty0 (6):\penalty0 131--151, 2023.

\bibitem[He et~al.(2016)He, Zhang, Ren, and Sun]{DBLP:conf/cvpr/HeZRS16}
Kaiming He, Xiangyu Zhang, Shaoqing Ren, and Jian Sun.
\newblock Deep residual learning for image recognition.
\newblock In \emph{{IEEE/CVF} Conference on Computer Vision and Pattern Recognition (CVPR)}, pages 770--778, 2016.

\bibitem[Hu et~al.(2022{\natexlab{a}})Hu, yelong shen, Wallis, Allen-Zhu, Li, Wang, Wang, and Chen]{DBLP:conf/iclr/HuSWALWWC22}
Edward~J Hu, yelong shen, Phillip Wallis, Zeyuan Allen-Zhu, Yuanzhi Li, Shean Wang, Lu Wang, and Weizhu Chen.
\newblock Lo{RA}: Low-rank adaptation of large language models.
\newblock In \emph{International Conference on Learning Representations (ICLR)}, 2022{\natexlab{a}}.

\bibitem[Hu et~al.(2022{\natexlab{b}})Hu, Fang, Lei, Zhong, and Chen]{huWhere2commCommunicationEfficientCollaborative2022a}
Yue Hu, Shaoheng Fang, Zixing Lei, Yiqi Zhong, and Siheng Chen.
\newblock Where2comm: Communication-efficient collaborative perception via spatial confidence maps.
\newblock In \emph{Advances in Neural Information Processing Systems (NeurIPS)}, pages 4874--4886, 2022{\natexlab{b}}.

\bibitem[Hu et~al.(2024)Hu, Peng, Liu, Ge, Liu, and Chen]{DBLP:conf/cvpr/HuPLGLC24}
Yue Hu, Juntong Peng, Sifei Liu, Junhao Ge, Si Liu, and Siheng Chen.
\newblock Communication-efficient collaborative perception via information filling with codebook.
\newblock In \emph{Proceedings of the IEEE/CVF Conference on Computer Vision and Pattern Recognition (CVPR)}, pages 15481--15490, 2024.

\bibitem[Jia et~al.(2022)Jia, Tang, Chen, Cardie, Belongie, Hariharan, and Lim]{jiaVisualPromptTuning2022}
Menglin Jia, Luming Tang, Bor{-}Chun Chen, Claire Cardie, Serge~J. Belongie, Bharath Hariharan, and Ser{-}Nam Lim.
\newblock Visual prompt tuning.
\newblock In \emph{European Conference on Computer Vision (ECCV)}, pages 709--727, 2022.

\bibitem[Lang et~al.(2019)Lang, Vora, Caesar, Zhou, Yang, and Beijbom]{DBLP:conf/cvpr/LangVCZYB19}
Alex~H. Lang, Sourabh Vora, Holger Caesar, Lubing Zhou, Jiong Yang, and Oscar Beijbom.
\newblock Pointpillars: Fast encoders for object detection from point clouds.
\newblock In \emph{Proceedings of the IEEE/CVF Conference on Computer Vision and Pattern Recognition (CVPR)}, 2019.

\bibitem[Lei et~al.(2022)Lei, Ren, Hu, Zhang, and Chen]{leiLatencyAwareCollaborativePerception2022}
Zixing Lei, Shunli Ren, Yue Hu, Wenjun Zhang, and Siheng Chen.
\newblock Latency-aware collaborative perception.
\newblock In \emph{European Conference on Computer Vision (ECCV)}, page 316–332, 2022.

\bibitem[Li et~al.(2024)Li, Li, Liu, Xu, Tu, Guo, Li, and Yu]{DBLP:journals/corr/abs-2403-11371}
Baolu Li, Jinlong Li, Xinyu Liu, Runsheng Xu, Zhengzhong Tu, Jiacheng Guo, Xiaopeng Li, and Hongkai Yu.
\newblock {V2X-DGW:} domain generalization for multi-agent perception under adverse weather conditions.
\newblock \emph{arXiv preprint arXiv:2403.11371}, 2024.

\bibitem[Li et~al.(2023)Li, Zhang, Yao, Song, Hao, Zhao, Li, and Chen]{liLearningDomainAwareDetection2023}
Haochen Li, Rui Zhang, Hantao Yao, Xinkai Song, Yifan Hao, Yongwei Zhao, Ling Li, and Yunji Chen.
\newblock Learning domain-aware detection head with prompt tuning.
\newblock In \emph{Advances in Neural Information Processing Systems (NeurIPS)}, pages 4248--4262, 2023.

\bibitem[Liu et~al.(2023)Liu, Gao, Chen, Peng, Kong, Wang, Xu, Jiang, Xiang, Ma, and Wang]{liuVehicleeverythingAutonomousDriving2023}
Si Liu, Chen Gao, Yuan Chen, Xingyu Peng, Xianghao Kong, Kun Wang, Runsheng Xu, Wentao Jiang, Hao Xiang, Jiaqi Ma, and Miao Wang.
\newblock Towards vehicle-to-everything autonomous driving: {A} survey on collaborative perception.
\newblock \emph{arXiv preprint arXiv:2308.16714}, 2023.

\bibitem[Liu et~al.(2020)Liu, Tian, Glaser, and Kira]{liuWhen2comMultiAgentPerception2020}
Yen-Cheng Liu, Junjiao Tian, Nathaniel Glaser, and Zsolt Kira.
\newblock When2com: Multi-agent perception via communication graph grouping.
\newblock In \emph{Proceedings of the IEEE/CVF Conference on Computer Vision and Pattern Recognition (CVPR)}, pages 4105--4114, 2020.

\bibitem[Lu et~al.(2023)Lu, Li, Liu, Dianati, Feng, Chen, and Wang]{luRobustCollaborative3D2023}
Yifan Lu, Quanhao Li, Baoan Liu, Mehrdad Dianati, Chen Feng, Siheng Chen, and Yanfeng Wang.
\newblock Robust collaborative 3d object detection in presence of pose errors.
\newblock In \emph{IEEE International Conference on Robotics and Automation (ICRA)}, pages 4812--4818, 2023.

\bibitem[Lu et~al.(2024)Lu, Hu, Zhong, Wang, Wang, and Chen]{luExtensibleFrameworkOpen2024}
Yifan Lu, Yue Hu, Yiqi Zhong, Dequan Wang, Yanfeng Wang, and Siheng Chen.
\newblock An extensible framework for open heterogeneous collaborative perception.
\newblock In \emph{International Conference on Learning Representationss (ICLR)}, 2024.

\bibitem[Luo et~al.(2022)Luo, Zhang, Yuan, and Li]{luoComplementarityEnhancedRedundancyMinimizedCollaboration2022}
Guiyang Luo, Hui Zhang, Quan Yuan, and Jinglin Li.
\newblock Complementarity-enhanced and redundancy-minimized collaboration network for multi-agent perception.
\newblock In \emph{Proceedings of the 30th ACM International Conference on Multimedia (ACM MM)}, page 3578–3586, 2022.

\bibitem[Luo et~al.(2025)Luo, Yuan, Luo, Xia, Yang, and Li]{luoPlugPlayRepresentation}
Tianyou Luo, Quan Yuan, Guiyang Luo, Yuchen Xia, Yujia Yang, and Jinglin Li.
\newblock Plug and play: A representation enhanced domain adapter for collaborative perception.
\newblock In \emph{European Conference on Computer Vision (ECCV)}, pages 287--303, 2025.

\bibitem[Ma et~al.(2023)Ma, Sun, Yang, and Yang]{DBLP:conf/cvpr/Ma0Y023}
Tianyi Ma, Yifan Sun, Zongxin Yang, and Yi Yang.
\newblock Prod: Prompting-to-disentangle domain knowledge for cross-domain few-shot image classification.
\newblock In \emph{Proceedings of the IEEE/CVF Conference on Computer Vision and Pattern Recognition (CVPR)}, pages 19754--19763, 2023.

\bibitem[Ren et~al.(2024)Ren, Lei, Wang, Dianati, Wang, Chen, and Zhang]{renInterruptionAwareCooperativePerception2024}
Shunli Ren, Zixing Lei, Zi Wang, Mehrdad Dianati, Yafei Wang, Siheng Chen, and Wenjun Zhang.
\newblock Interruption-aware cooperative perception for {V2X} communication-aided autonomous driving.
\newblock \emph{{IEEE} Trans. Intell. Veh.}, 9\penalty0 (4):\penalty0 4698--4714, 2024.

\bibitem[Shao et~al.(2025)Shao, Luo, Yuan, Chen, Liu, Gong, and Li]{shaoHetecooperFeatureCollaboration}
Congzhang Shao, Guiyang Luo, Quan Yuan, Yifu Chen, Yilin Liu, Kexin Gong, and Jinglin Li.
\newblock Hetecooper: Feature collaboration graph for heterogeneous collaborative perception.
\newblock In \emph{European Conference on Computer Vision (ECCV)}, pages 162--178, 2025.

\bibitem[Singha et~al.(2023)Singha, Pal, Jha, and Banerjee]{singhaADCLIPAdaptingDomains2023}
Mainak Singha, Harsh Pal, Ankit Jha, and Biplab Banerjee.
\newblock Ad-clip: Adapting domains in prompt space using clip.
\newblock In \emph{2023 IEEE/CVF International Conference on Computer Vision Workshops (ICCVW)}, pages 4357--4366, 2023.

\bibitem[Tan and Le(2019)]{DBLP:conf/icml/TanL19}
Mingxing Tan and Quoc~V. Le.
\newblock Efficientnet: Rethinking model scaling for convolutional neural networks.
\newblock In \emph{Proceedings of Machine Learning Research (ICML)}, pages 6105--6114, 2019.

\bibitem[Wang et~al.(2023)Wang, Zeng, and Gong]{DBLP:journals/tits/WangZG23}
Junyong Wang, Yuan Zeng, and Yi Gong.
\newblock Collaborative 3d object detection for autonomous vehicles via learnable communications.
\newblock \emph{{IEEE} Trans. Intell. Transp. Syst.}, 24\penalty0 (9):\penalty0 9804--9816, 2023.

\bibitem[Woo et~al.(2018)Woo, Park, Lee, and Kweon]{CBAM}
Sanghyun Woo, Jongchan Park, Joon-Young Lee, and In~So Kweon.
\newblock Cbam: Convolutional block attention module.
\newblock In \emph{European Conference on Computer Vision (ECCV)}, page 3–19, 2018.

\bibitem[Xiang et~al.(2023)Xiang, Xu, and Ma]{xiangHMViTHeteromodalVehicleVehicle2023}
Hao Xiang, Runsheng Xu, and Jiaqi Ma.
\newblock Hm-vit: Hetero-modal vehicle-to-vehicle cooperative perception with vision transformer.
\newblock In \emph{Proceedings of the IEEE/CVF International Conference on Computer Vision (ICCV)}, pages 284--295, 2023.

\bibitem[Xu et~al.(2022{\natexlab{a}})Xu, Xiang, Tu, Xia, Yang, and Ma]{xuV2XViTVehicleEverythingCooperative2022}
Runsheng Xu, Hao Xiang, Zhengzhong Tu, Xin Xia, Ming{-}Hsuan Yang, and Jiaqi Ma.
\newblock V2x-vit: Vehicle-to-everything cooperative perception with vision transformer.
\newblock In \emph{European Conference on Computer Vision (ECCV)}, pages 107--124, 2022{\natexlab{a}}.

\bibitem[Xu et~al.(2022{\natexlab{b}})Xu, Xiang, Xia, Han, Li, and Ma]{DBLP:conf/icra/XuXXHLM22}
Runsheng Xu, Hao Xiang, Xin Xia, Xu Han, Jinlong Li, and Jiaqi Ma.
\newblock {OPV2V:} an open benchmark dataset and fusion pipeline for perception with vehicle-to-vehicle communication.
\newblock In \emph{2022 IEEE International Conference on Robotics and Automation (ICRA)}, pages 2583--2589, 2022{\natexlab{b}}.

\bibitem[Xu et~al.(2023{\natexlab{a}})Xu, Li, Dong, Yu, and Ma]{xuBridgingDomainGap2023}
Runsheng Xu, Jinlong Li, Xiaoyu Dong, Hongkai Yu, and Jiaqi Ma.
\newblock Bridging the {{domain gap}} for {{multi-agent perception}}.
\newblock In \emph{2023 {{IEEE International Conference}} on {{Robotics}} and {{Automation}} ({{ICRA}})}, pages 6035--6042, 2023{\natexlab{a}}.

\bibitem[Xu et~al.(2023{\natexlab{b}})Xu, Tu, Xiang, Shao, Zhou, and Ma]{xuCoBEVTCooperativeBirds2022}
Runsheng Xu, Zhengzhong Tu, Hao Xiang, Wei Shao, Bolei Zhou, and Jiaqi Ma.
\newblock Cobevt: Cooperative bird’s eye view semantic segmentation with sparse transformers.
\newblock In \emph{Proceedings of The 6th Conference on Robot Learning (CoRL)}, pages 989--1000, 2023{\natexlab{b}}.

\bibitem[Yan et~al.(2018)Yan, Mao, and Li]{DBLP:journals/sensors/YanML18}
Yan Yan, Yuxing Mao, and Bo Li.
\newblock {SECOND:} sparsely embedded convolutional detection.
\newblock \emph{Sensors}, 18\penalty0 (10):\penalty0 3337, 2018.

\bibitem[Yang et~al.(2023)Yang, Yang, Zhang, Li, Liu, Liu, Wang, Sun, and Song]{Yang_2023_ICCV}
Kun Yang, Dingkang Yang, Jingyu Zhang, Mingcheng Li, Yang Liu, Jing Liu, Hanqi Wang, Peng Sun, and Liang Song.
\newblock Spatio-temporal domain awareness for multi-agent collaborative perception.
\newblock In \emph{Proceedings of the IEEE/CVF International Conference on Computer Vision (ICCV)}, pages 23383--23392, 2023.

\bibitem[Yang et~al.(2024)Yang, Wu, Liu, Li, Zhang, Pan, Gan, Chen, and Zhang]{yangExploringSparseVisual2024}
Senqiao Yang, Jiarui Wu, Jiaming Liu, Xiaoqi Li, Qizhe Zhang, Mingjie Pan, Yulu Gan, Zehui Chen, and Shanghang Zhang.
\newblock Exploring {{sparse visual prompt}} for {{domain adaptive dense prediction}}.
\newblock \emph{In AAAI Conference on Artificial Intelligence (AAAI)}, pages 16334--16342, 2024.

\bibitem[Yin et~al.(2024)Yin, Tian, Lin, Duan, Zhou, Zhao, and Cao]{10265751}
Hongbo Yin, Daxin Tian, Chunmian Lin, Xuting Duan, Jianshan Zhou, Dezong Zhao, and Dongpu Cao.
\newblock V2vformer++: Multi-modal vehicle-to-vehicle cooperative perception via global-local transformer.
\newblock \emph{{IEEE} Trans. Intell. Transp. Syst.}, 25\penalty0 (2):\penalty0 2153--2166, 2024.

\bibitem[Yu et~al.(2022)Yu, Luo, Shu, Huo, Yang, Shi, Guo, Li, Hu, Yuan, and Nie]{DBLP:conf/cvpr/YuLSHYSGLHYN22}
Haibao Yu, Yizhen Luo, Mao Shu, Yiyi Huo, Zebang Yang, Yifeng Shi, Zhenglong Guo, Hanyu Li, Xing Hu, Jirui Yuan, and Zaiqing Nie.
\newblock {DAIR-V2X:} {A} large-scale dataset for vehicle-infrastructure cooperative 3d object detection.
\newblock In \emph{{IEEE/CVF} Conference on Computer Vision and Pattern Recognition (CVPR)}, pages 21329--21338, 2022.

\bibitem[Zhao et~al.(2023)Zhao, Zhang, and Zou]{zhaoBM2CPEfficientCollaborative2023}
Binyu Zhao, Wei Zhang, and Zhaonian Zou.
\newblock Bm2cp: Efficient collaborative perception with lidar-camera modalities.
\newblock In \emph{Proceedings of The 7th Conference on Robot Learning (CoRL)}, pages 1022--1035, 2023.

\bibitem[Zhou and Tuzel(2018)]{DBLP:conf/cvpr/ZhouT18}
Yin Zhou and Oncel Tuzel.
\newblock Voxelnet: End-to-end learning for point cloud based 3d object detection.
\newblock In \emph{Proceedings of the IEEE/CVF Conference on Computer Vision and Pattern Recognition (CVPR)}, pages 4490--4499, 2018.

\end{thebibliography}
}

\setcounter{section}{0}
\setcounter{table}{0}
\setcounter{figure}{0}
\clearpage
\setcounter{page}{1}
\maketitlesupplementary

\section{Experimental Details}

When adapting to a new neighbor agent, trainable parameters include a specific prompt for the neighbor agent and a resizer to align the size of the neighbor features with the ego features. The resizer consists of a max-pooling layer and a $1\times1$ convolution. Specifically, if the ego feature size is $C_1\times H_1\times W_1$ and the neighbor feature size is $C_2\times H_2\times W_2$, the number of parameters for the specific prompt is $C_2\times H_1\times W_1$, and the number of parameters for the resizer is $C_1\times C_2$. The trainable parameter numbers for adapting to new neighbor agents with pp8 as the ego agent are shown in Table \ref{tab:paramnum}.

\begin{table}[b]
  \centering
  \renewcommand\arraystretch{1.1}
  \resizebox{0.48\textwidth}{!}{
  \begin{tabular}{@{}>{\raggedright\arraybackslash}p{0.3cm}>{\raggedright\arraybackslash}p{0.8cm}c>{\centering\arraybackslash}p{2.3cm}>{\centering\arraybackslash}p{0.9cm}>{\centering\arraybackslash}p{1.8cm}@{}}
    \toprule
     & & & \multicolumn{3}{c}{ Parameters (M)} \\
    \cmidrule[\heavyrulewidth]{4-6}
    Encoder & & Feature Size & Specific Prompt & Resizer & Total \\
    \midrule[\heavyrulewidth]
    \makebox[1cm][l]{pp8} & \multirow{3}{*}{\cite{DBLP:conf/cvpr/LangVCZYB19}} & $256\times 50\times 176$ & & & \\
    pp6 & & $384\times 64\times 256$ & 3.38 & 0.10 & 3.48 \\
    pp4 & & $384\times 100\times 352$ & 3.38 & 0.10 & 3.48 \\
    \midrule
    vn6 & \multirow{2}{*}{\cite{DBLP:journals/sensors/YanML18}} & $128\times 128\times 512$ & 1.13 & 0.03 & 1.16 \\
    vn4 & & $128\times 200\times 704$ & 1.13 & 0.03 & 1.16 \\
    \midrule
    sd2 & \multirow{2}{*}{\cite{DBLP:conf/cvpr/ZhouT18}} & $512\times 50\times 176$ & 4.51 & 0.13 & 4.64 \\
    sd1 & & $512\times 100\times 352$ & 4.51 & 0.13 & 4.64 \\
    \bottomrule[\heavyrulewidth]
  \end{tabular}
  }
  \caption{Trainable parameter numbers of different encoders.}
  \label{tab:paramnum}
\end{table}

\section{Additional Ablation Study}

We perform ablation experiments on three loss components: style loss (regulating both shared and agent-specific semantics), adversary loss (regulating shared semantics), and single loss (regulating agent-specific semantics), with results detailed in Table \ref{loss}, following the main paper's settings.

\begin{table}[b]
    \centering
    \resizebox{0.45\textwidth}{!}{
        \begin{tabular}{@{}lccc@{}}
            \toprule
             & \textbf{pp8-pp4} & \textbf{pp8-sd1} & \textbf{pp8-vn6} \\
            \midrule[\heavyrulewidth]
            PolyInter & \textbf{77.2 / 65.8} & \textbf{79.4 / 66.3} & \textbf{72.8 / 55.1} \\
            \quad -w/o style & 75.2 / 62.2 & 77.2 / 63.5 & 69.6 / 50.1 \\
            \quad -w/o adversary & 77.0 / 63.4 & 79.1 / 63.9 & 71.8 / 54.0 \\
            \quad -w/o single & 76.4 / 59.7 & 78.4 / 61.5 & 72.2 / 52.0 \\
            \bottomrule[\heavyrulewidth]
        \end{tabular}
    }
    \caption{Ablation study of loss functions.}
    \label{loss}
\end{table}

\section{Performance of Multi-Modal Fusion}

We conduct multi-modal fusion experiments by integrating LiDAR and camera data, using two image encoders, EfficientNet \cite{DBLP:conf/icml/TanL19} and ResNet \cite{DBLP:conf/cvpr/HeZRS16}, and conduct two sets of experiments. In the first set, the base model is trained with the combination of pp8-pp4-EfficientNet, where the ego agent collaborates with new neighbor agents using ResNet in Generalization Phase. In the second set, the base model is trained with pp8-vn4-ResNet, and the ego agent collaborates with new neighbor agents using EfficientNet in Generalization Phase. As shown in Table \ref{modal}, PolyInter performs well in multi-modal fusion.

\begin{table*}[h]
  \centering
  \resizebox{0.88\textwidth}{!}{
  \begin{tabular}{@{}>{\raggedright\arraybackslash}p{3.2cm}cccccc@{}}
    \toprule
     & \multicolumn{3}{c}{{w/ F-cooper Fusion}} & \multicolumn{3}{c}{{w/ CoBEVT Fusion}}\\
    \cmidrule[\heavyrulewidth](l{7pt}r{7pt}){2-4} \cmidrule[\heavyrulewidth](l{7pt}r{7pt}){5-7}
     & \textbf{Ours} & \textbf{PnPDA} \cite{luoPlugPlayRepresentation} & \textbf{MPDA} \cite{xuBridgingDomainGap2023} & \textbf{Ours} & \textbf{PnPDA} \cite{luoPlugPlayRepresentation} & \textbf{MPDA} \cite{xuBridgingDomainGap2023}\\
    \midrule
    \makebox[2cm][l]{pp8-ResNet \cite{DBLP:conf/cvpr/HeZRS16}}& \textbf{72.6 / 56.9} & 61.9 / 42.5 & 70.7 / 56.6 & \textbf{75.4 / 57.9} & 68.4 / 46.6 & 71.4 / 56.5\\
    \makebox[2cm][l]{pp8-EfficientNet \cite{DBLP:conf/icml/TanL19}}& \textbf{72.2 / 57.3} & 60.1 / 41.0 & 67.4 / 54.4 & \textbf{72.6 / 57.5} & 67.8 / 46.8 & 70.2 / 56.3\\
    \bottomrule
  \end{tabular}
  }
  \caption{LiDAR + camera performance in AP@0.5/AP@0.7.}
  \label{modal}
\end{table*}

\section{New Datasets}

\subsection{Performance Comparison on V2XSet}

An open dataset, V2XSet \cite{xuV2XViTVehicleEverythingCooperative2022}, is used in the comparative experiments. Compared to the OPV2V dataset, V2XSet incorporates vehicle-to-everything cooperation and realistic noise simulation. The experimental results comparing PolyInter with PnPDA \cite{luoPlugPlayRepresentation} and MPDA \cite{xuBridgingDomainGap2023} are presented in Table \ref{v2xset}.

\subsection{Performance Comparison on DAIR-V2X}

We also conduct experiments on the real-world DAIR-V2X \cite{DBLP:conf/cvpr/YuLSHYSGLHYN22} dataset. We use pp8-pp4-vn4 combination in the base model training phase. The experimental results comparing PolyInter with PnPDA \cite{luoPlugPlayRepresentation} and MPDA \cite{xuBridgingDomainGap2023} are presented in Table \ref{dairv2x}.

\begin{table*}[h!]
  \centering
  \resizebox{0.97\textwidth}{!}{
  \begin{tabular}{@{}>{\raggedright\arraybackslash}p{1.1cm}lcccccccc@{}}
    \toprule
    \multicolumn{2}{c}{\multirow{2}{*}{\diagbox[width=2.5cm,height=1cm]{Scenarios }{Interpreter}}} & \multicolumn{3}{c}{{w/ F-cooper \cite{DBLP:conf/edge/ChenMTGYF19} Fusion}}& & \multicolumn{3}{c}{{w/ CoBEVT \cite{xuCoBEVTCooperativeBirds2022} Fusion}}  \\
    \cmidrule[\heavyrulewidth]{3-5} \cmidrule[\heavyrulewidth]{7-9}
     & & \textbf{PolyInter (Ours)} & \textbf{PnPDA} \cite{luoPlugPlayRepresentation} & \textbf{MPDA} \cite{xuBridgingDomainGap2023} &   & \textbf{PolyInter (Ours)} & \textbf{PnPDA} \cite{luoPlugPlayRepresentation} & \textbf{MPDA} \cite{xuBridgingDomainGap2023} \\
    \midrule
    \makebox[2cm][l]{pp8-pp4*}&\multirow{2}{*}{\cite{DBLP:conf/cvpr/LangVCZYB19}} & \textbf{84.1 / 71.3} & \multirow{2}{*}{80.6 / 55.0} & \multirow{2}{*}{77.8 / 55.4} &    & \textbf{86.4 / 72.2} & \multirow{2}{*}{84.5 / 63.9} & \multirow{2}{*}{84.2 / 70.5} \\
    \makebox[2cm][l]{pp8-pp4+} &  & \textbf{86.7 / 72.2} & &    & & \textbf{86.5 / 72.1} & & \\
    \midrule
    \makebox[2cm][l]{pp8-sd1*}&\multirow{2}{*}{\cite{DBLP:conf/cvpr/LangVCZYB19,DBLP:journals/sensors/YanML18}} & \textbf{85.8 / 70.3} & \multirow{2}{*}{83.5 / 63.6} & \multirow{2}{*}{78.4 / 54.0} &    & \textbf{87.7 / 76.8} & \multirow{2}{*}{86.9 / 63.3} & \multirow{2}{*}{81.4 / 66.6} \\
    \makebox[2cm][l]{pp8-sd1+} &  & \textbf{87.9 / 74.8} & &    & & \textbf{87.4 / 74.5} & & \\
    \midrule
    \makebox[2cm][l]{pp8-vn6*}&\multirow{2}{*}{\cite{DBLP:conf/cvpr/LangVCZYB19,DBLP:conf/cvpr/ZhouT18}} & \textbf{80.5 / 71.3} & \multirow{2}{*}{75.9 / 51.7} & \multirow{2}{*}{69.5 / 50.7} &    & \textbf{83.8 / 65.4} & \multirow{2}{*}{79.7 / 51.1} & \multirow{2}{*}{70.7 / 51.7} \\
    \makebox[2cm][l]{pp8-vn6+} &  & \textbf{84.0 / 62.9} & &    & & \textbf{83.7 / 63.6} & & \\
    
    \bottomrule
  \end{tabular}
  }
  \caption{Comparison with PnPDA and MPDA on V2XSet dataset.}
  \label{v2xset}
\end{table*}

\begin{table*}[h!]
  \centering
  \resizebox{0.97\textwidth}{!}{
  \begin{tabular}{@{}>{\raggedright\arraybackslash}p{1.1cm}lcccccccc@{}}
    \toprule
    \multicolumn{2}{c}{\multirow{2}{*}{\diagbox[width=2.5cm,height=1cm]{Scenarios }{Interpreter}}} & \multicolumn{3}{c}{{w/ F-cooper \cite{DBLP:conf/edge/ChenMTGYF19} Fusion}}& & \multicolumn{3}{c}{{w/ CoBEVT \cite{xuCoBEVTCooperativeBirds2022} Fusion}}  \\
    \cmidrule[\heavyrulewidth]{3-5} \cmidrule[\heavyrulewidth]{7-9}
     & & \textbf{PolyInter (Ours)} & \textbf{PnPDA} \cite{luoPlugPlayRepresentation} & \textbf{MPDA} \cite{xuBridgingDomainGap2023} &   & \textbf{PolyInter (Ours)} & \textbf{PnPDA} \cite{luoPlugPlayRepresentation} & \textbf{MPDA} \cite{xuBridgingDomainGap2023} \\
    \midrule
    \makebox[2cm][l]{pp8-pp4 \cite{DBLP:conf/cvpr/LangVCZYB19}}  & & \textbf{65.2 / 38.5} & 59.3 / 33.9 & 64.8 / 32.1& & \textbf{66.8 / 39.9} & 62.9 / 34.4 & 65.0 / 35.1 \\
    \midrule
    \makebox[2cm][l]{pp8-sd1 \cite{DBLP:conf/cvpr/LangVCZYB19,DBLP:journals/sensors/YanML18}}& & \textbf{65.0 / 38.4} & 63.6 / 32.8 & 64.5 / 34.2& & \textbf{67.5 / 40.0} & 63.4 / 36.0 & 65.5 / 35.3 \\
    \midrule
    \makebox[2cm][l]{pp8-vn6 \cite{DBLP:conf/cvpr/LangVCZYB19,DBLP:conf/cvpr/ZhouT18}}& & \textbf{63.9 / 38.2} & 49.1 / 30.0 & 62.8 / 33.6& & \textbf{65.9 / 39.6} & 63.9 / 34.1 & 64.8 / 34.6 \\
    
    \bottomrule
  \end{tabular}
  }
  \caption{Comparison with PnPDA and MPDA on DAIR-V2X dataset.}
  \label{dairv2x}
\end{table*}

\section{Additional Experiments}

\subsection{Three-agent Collaborative Perception}
We compare the performance of PolyInter with PnPDA \cite{luoPlugPlayRepresentation} and MPDA \cite{xuBridgingDomainGap2023} in the immutable heterogeneous scenario of three-agent collaborative perception, as shown in Table \ref{three-agent}. The three collaborating agents are set as three different agent types, in the format of ``ego-neb1-neb2". The selected scenarios include pp8-pp4-vn6, pp8-pp4-sd1, and pp8-vn6-sd1. In three-agent collaboration, the ego agent's interpreter separately interprets the heterogeneous features of the two neighbor agents into the ego agent's semantic space. The interpreted neighbor features, combined with the ego features, are fed into the fusion module and the detection head on the ego agent to produce the collaborative perception results. The remaining settings are consistent with those described in \cref{subsec:Performance}.

\begin{table*}[h!]
  \centering
  \resizebox{1\textwidth}{!}{
  \begin{tabular}{@{}>{\raggedright\arraybackslash}p{1.7cm}lcccccccc@{}}
    \toprule
    \multicolumn{2}{c}{\multirow{2}{*}{\diagbox[width=2.5cm,height=1cm]{Scenarios }{Interpreter}}} & \multicolumn{3}{c}{{w/ F-cooper \cite{DBLP:conf/edge/ChenMTGYF19} Fusion}}& & \multicolumn{3}{c}{{w/ CoBEVT \cite{xuCoBEVTCooperativeBirds2022} Fusion}}  \\
    \cmidrule[\heavyrulewidth]{3-5} \cmidrule[\heavyrulewidth]{7-9}
     & & \textbf{PolyInter (Ours)} & \textbf{PnPDA} \cite{luoPlugPlayRepresentation} & \textbf{MPDA} \cite{xuBridgingDomainGap2023} &   & \textbf{PolyInter (Ours)} & \textbf{PnPDA} \cite{luoPlugPlayRepresentation} & \textbf{MPDA} \cite{xuBridgingDomainGap2023} \\
    \midrule
    \makebox[2cm][l]{pp8-pp4-vn6*}&\multirow{2}{*}{\cite{DBLP:conf/cvpr/LangVCZYB19,DBLP:conf/cvpr/ZhouT18}} & \textbf{77.0 / 61.4} & \multirow{2}{*}{63.5 / 46.4} & \multirow{2}{*}{69.1 / 49.5} &    & \textbf{79.2 / 68.6} & \multirow{2}{*}{78.0 / 62.5} & \multirow{2}{*}{72.9 / 57.2} \\
    \makebox[2cm][l]{pp8-pp4-vn6+} &  & \textbf{78.0 / 66.4} & &    & & \textbf{78.2 / 67.0} & & \\
    \midrule
    \makebox[2cm][l]{pp8-pp4-sd1*}&\multirow{2}{*}{\cite{DBLP:conf/cvpr/LangVCZYB19,DBLP:journals/sensors/YanML18}} & \textbf{80.3 / 67.2} & \multirow{2}{*}{74.5 / 41.8} & \multirow{2}{*}{73.2 / 45.8} &    & \textbf{83.5 / 74.5} & \multirow{2}{*}{79.1 / 65.3} & \multirow{2}{*}{79.6 / 67.0} \\
    \makebox[2cm][l]{pp8-pp4-sd1+} &  & \textbf{79.1 / 69.1} & &    & & \textbf{81.1 / 70.9} & & \\
    \midrule
    \makebox[2cm][l]{pp8-vn6-sd1*}&\multirow{2}{*}{\cite{DBLP:conf/cvpr/LangVCZYB19,DBLP:conf/cvpr/ZhouT18,DBLP:journals/sensors/YanML18}} & \textbf{79.4 / 65.9} & \multirow{2}{*}{68.0 / 48.5} & \multirow{2}{*}{61.4 / 44.9} &    & \textbf{80.2 / 70.7} & \multirow{2}{*}{72.0 / 55.2} & \multirow{2}{*}{71.5 / 50.9} \\
    \makebox[2cm][l]{pp8-vn6-sd1+} &  & \textbf{78.8 / 65.9} & &    & & \textbf{78.6 / 68.3} & & \\
    
    \bottomrule
  \end{tabular}
  }
  \caption{Comparison with PnPDA and MPDA for three-agent collaborative perception. Our experiments include three heterogeneous scenarios (in the format of ``ego-neb1-neb2"): pp8-pp4-vn6, pp8-pp4-sd1, and pp8-vn6-sd1.}
  \label{three-agent}
\end{table*}

\subsection{Two-Stage Interpretation of PnPDA}

PnPDA \cite{luoPlugPlayRepresentation} adopts a two-stage strategy, where in practical applications, the neighbor features are first interpreted into a standard semantic space and then further interpreted into the ego agent's semantic space. The results presented in \cref{subsec:Performance} are from one-stage interpretation, where neighbor features are directly interpreted into the ego agent's semantic space without passing through the standard semantic space. The performance of two-stage interpretation of PnPDA is shown in Table \ref{tab:pnpda2stage}. With pp8 as the ego agent and pp4, sd1, and vn6 as the neighbor agents, two agent types, pp4 and vn4, are used as standard semantic spaces, consistent with the settings in \cite{luoPlugPlayRepresentation}. The two-stage interpretation, by passing through the standard semantic space, incurs two stages of semantic loss, which considerably diminishes the collaborative performance.

\begin{table}[h!]
  \centering
  \renewcommand\arraystretch{1.1}
  \resizebox{0.48\textwidth}{!}{
  \begin{tabular}{@{}l|c|c|c|c@{}}
    \toprule
    Standard Semantic & \multicolumn{2}{c|}{pp4 \cite{DBLP:conf/cvpr/LangVCZYB19}} & \multicolumn{2}{c}{vn4 \cite{DBLP:journals/sensors/YanML18}} \\
    \midrule[\heavyrulewidth]
    Fusion Method & F-cooper \cite{DBLP:conf/edge/ChenMTGYF19} & CoBEVT \cite{xuCoBEVTCooperativeBirds2022} & F-cooper \cite{DBLP:conf/edge/ChenMTGYF19} & CoBEVT \cite{xuCoBEVTCooperativeBirds2022} \\
    \midrule
    pp8-pp4 \cite{DBLP:conf/cvpr/LangVCZYB19} & 77.1 / 51.1 & 79.2 / 62.3 & 69.1 / 50.8 & 73.9 / 57.5 \\
    pp8-sd1 \cite{DBLP:conf/cvpr/LangVCZYB19,DBLP:conf/cvpr/ZhouT18} & 58.7 / 35.0 & 63.3 / 42.1 & 63.9 / 41.5 & 65.4 / 49.7 \\
    pp8-vn6 \cite{DBLP:conf/cvpr/LangVCZYB19,DBLP:journals/sensors/YanML18} & 43.6 / 24.8 & 57.5 / 30.2 & 50.8 / 31.1 & 59.1 / 34.7 \\
    \bottomrule[\heavyrulewidth]
  \end{tabular}
  }
  \caption{Two-stage interpretation performance of PnPDA.}
  \label{tab:pnpda2stage}
\end{table}


\begin{table}[t]
  \centering
  \renewcommand\arraystretch{1.1}
  \resizebox{0.45\textwidth}{!}{
  \begin{tabular}{@{}c|cc|cc@{}}
    \toprule
    Fusion &  & Combination 1 &  & Combination 2 \\
    \midrule[\heavyrulewidth]
    \multirow{2}{*}{F-cooper \cite{DBLP:conf/edge/ChenMTGYF19}} & pp8-vn4 & 74.3 / 58.6 & pp8-pp4 & 77.2 / 65.6 \\
     & pp8-sd2 & 81.0 / 66.3 & pp8-vn4 & 74.1 / 61.0 \\
    \midrule[\heavyrulewidth]
    \multirow{2}{*}{CoBEVT \cite{xuCoBEVTCooperativeBirds2022}} & pp8-vn4 & 80.2 / 66.0 & pp8-pp4 & 80.2 / 67.0 \\
     & pp8-sd2 & 83.2 / 71.8 & pp8-vn4 & 78.0 / 63.4 \\     
    \bottomrule[\heavyrulewidth]
  \end{tabular}
  }
  \caption{Performance of PolyInter in phase \uppercase\expandafter{\romannumeral1}. Combination 1 consists of pp8-vn4-sd2, and Combination 2 consists of pp8-pp4-vn4.}
  \label{tab:phase1}
\end{table}

\begin{figure*}[t]
  \centering
   \includegraphics[width=\linewidth]{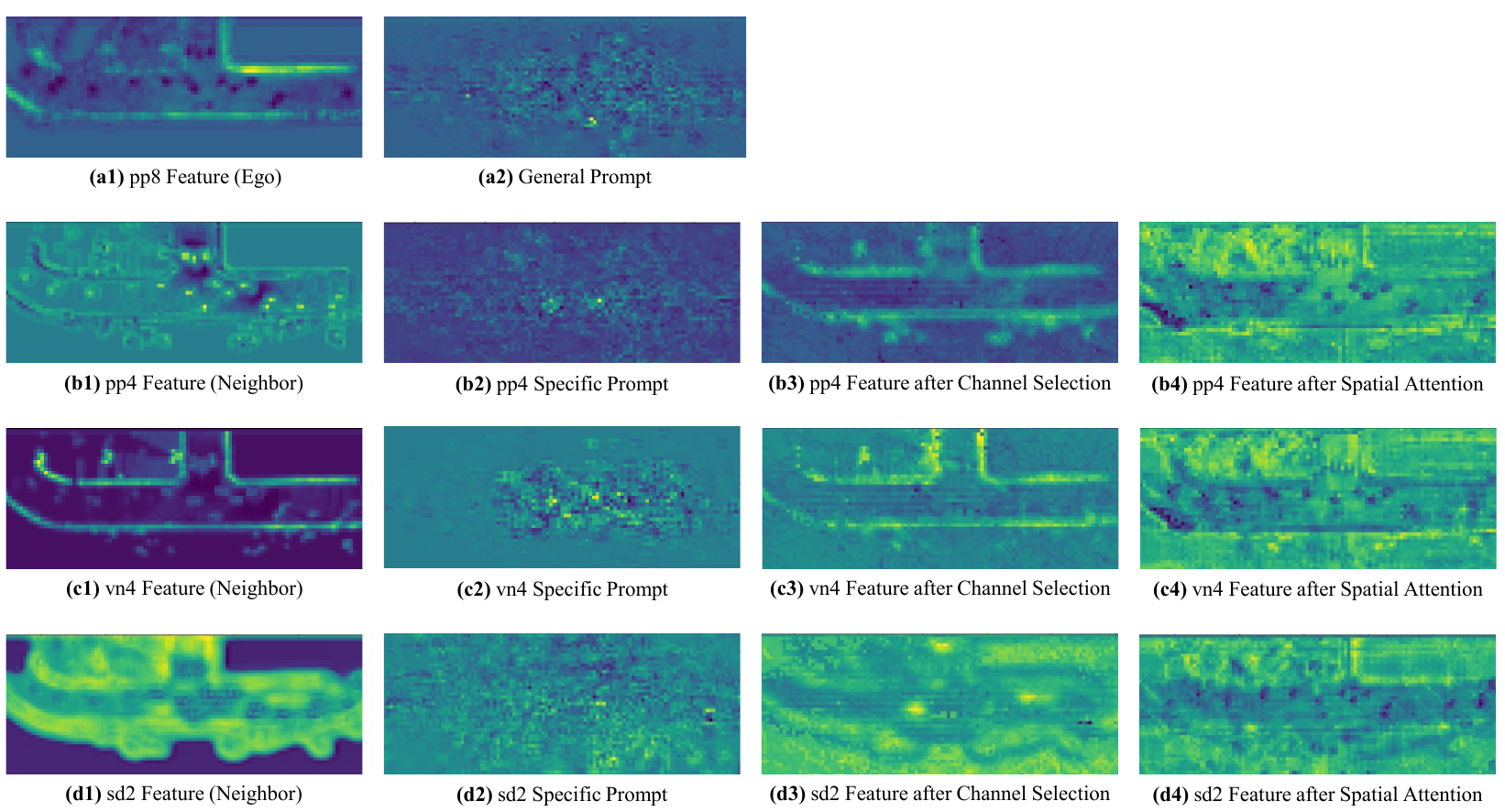}

   \caption{Visualization of the ego feature, the general prompt, the specific prompts corresponding to different neighbor agents, and the process of interpreting neighbor features into the ego agent's semantic space.}
   \label{additionalvis}
\end{figure*}

\subsection{Performance of PolyInter in Phase \uppercase\expandafter{\romannumeral1}}

The base model is trained with two different encoder combinations in phase \uppercase\expandafter{\romannumeral1}, in the format of "ego-neb1-neb2," including pp8-vn4-sd2 and pp8-pp4-vn4. The performance of the PolyInter base model under these settings is validated, with results shown in Table \ref{tab:phase1}.

\section{Additional Qualitative Evaluation}

As shown in Figure \ref{additionalvis}, the specific prompts and features for different neighbor agents are visualized. Taking pp8 as the ego agent, pp4, vn4, and sd1 were sequentially selected as neighbor agents. Features of different heterogeneous neighbor agents are matched with distinct specific prompts. The Channel Selection Module reorganizes neighbor features to align with the ego features, while the Spatial Attention Module establishes spatial connections. Finally, all heterogeneous 
neighbor features are interpreted into the ego agent's semantic space.



\end{document}